\documentclass[lettersize,journal]{IEEEtran}
\usepackage{amsmath,amsfonts}
\usepackage{algorithmic}
\usepackage{algorithm}
\usepackage{array}
\usepackage[caption=false,font=normalsize,labelfont=sf,textfont=sf]{subfig}
\usepackage{textcomp}
\usepackage{stfloats}
\usepackage{url}
\usepackage{verbatim}
\usepackage{graphicx}
\usepackage{cite}
\usepackage{multirow}
\usepackage{algorithm}
\usepackage{algorithmic}
\usepackage{booktabs}
\usepackage{xcolor}

\begin{document}

\title{Revisiting the Adversarial Robustness of Vision Language Models: a Multimodal Perspective}


\author{Wanqi Zhou, Shuanghao Bai, Danilo P. Mandic,~\IEEEmembership{Fellow,~IEEE,} Qibin Zhao, Badong Chen,~\IEEEmembership{Senior Member,~IEEE}
\thanks{This work was supported by the National Natural Science Foundation of China (grant numbers U21A20485, 62088102, and 62071132).
(Wanqi Zhou and Shuanghao Bai contributed equally to this work.) Corresponding authors: Badong Chen; Qibin Zhao. \\
\indent Wanqi Zhou, Shuanghao Bai, and Badong Chen are with the National Key Laboratory of Human-Machine Hybrid Augmented Intelligence, National Engineering Research Center for Visual Information and Applications, and Institute of Artificial Intelligence and Robotics, Xi’an Jiaotong University, Xi’an 710049, China (e-mail: zwq785915792@stu.xjtu.edu.cn; baishuanghao@stu.xjtu.edu.cn; chenbd@mail.xjtu.edu.cn).
Qibin Zhao is with RIKEN AIP, Tokyo, Japan, and Guangdong University of Technology, Guangdong 510006, China (e-mail: qibin.zhao@riken.jp).
Danilo P. Mandic is with the Department of Electrical Engineering, Imperial College London, London SW7 2BX, UK (e-mail: d.mandic@imperial.ac.uk).}
}


\markboth{Journal of \LaTeX\ Class Files,~Vol.~14, No.~8, August~2021}%
{Shell \MakeLowercase{\textit{et al.}}: A Sample Article Using IEEEtran.cls for IEEE Journals}

\IEEEpubid{\hspace{-15cm} \copyright~2024 IEEE}


\maketitle
\IEEEpubidadjcol

\begin{abstract}
Pretrained vision-language models (VLMs) like CLIP exhibit exceptional generalization across diverse downstream tasks. 
While recent studies reveal their vulnerability to adversarial attacks, research to date has primarily focused on enhancing the robustness of image encoders against image-based attacks, with defenses against text-based and multimodal attacks remaining largely unexplored. 
To this end, this work presents the first comprehensive study on improving the adversarial robustness of VLMs against attacks targeting image, text, and multimodal inputs.
This is achieved by proposing \textbf{m}ulti\textbf{m}odal \textbf{co}ntrastive \textbf{a}dversarial training (MMCoA).
Such an approach strengthens the robustness of both image and text encoders by aligning the clean text embeddings with adversarial image embeddings, and adversarial text embeddings with clean image embeddings.
The robustness of the proposed MMCoA is examined against existing defense methods over image, text, and multimodal attacks on the CLIP model.
Extensive experiments on 15 datasets across two tasks reveal the characteristics of different adversarial defense methods under distinct distribution shifts and dataset complexities across the three attack types. 
This paves the way for a unified framework of adversarial robustness against different modality attacks, opening up new possibilities for securing VLMs against multimodal attacks.
The code is available at https://github.com/ElleZWQ/MMCoA.git.
\end{abstract}


\section{Introduction}
\label{sec:intro}

\begin{figure}[ht]
  \centering
  \includegraphics[width=0.47\textwidth]{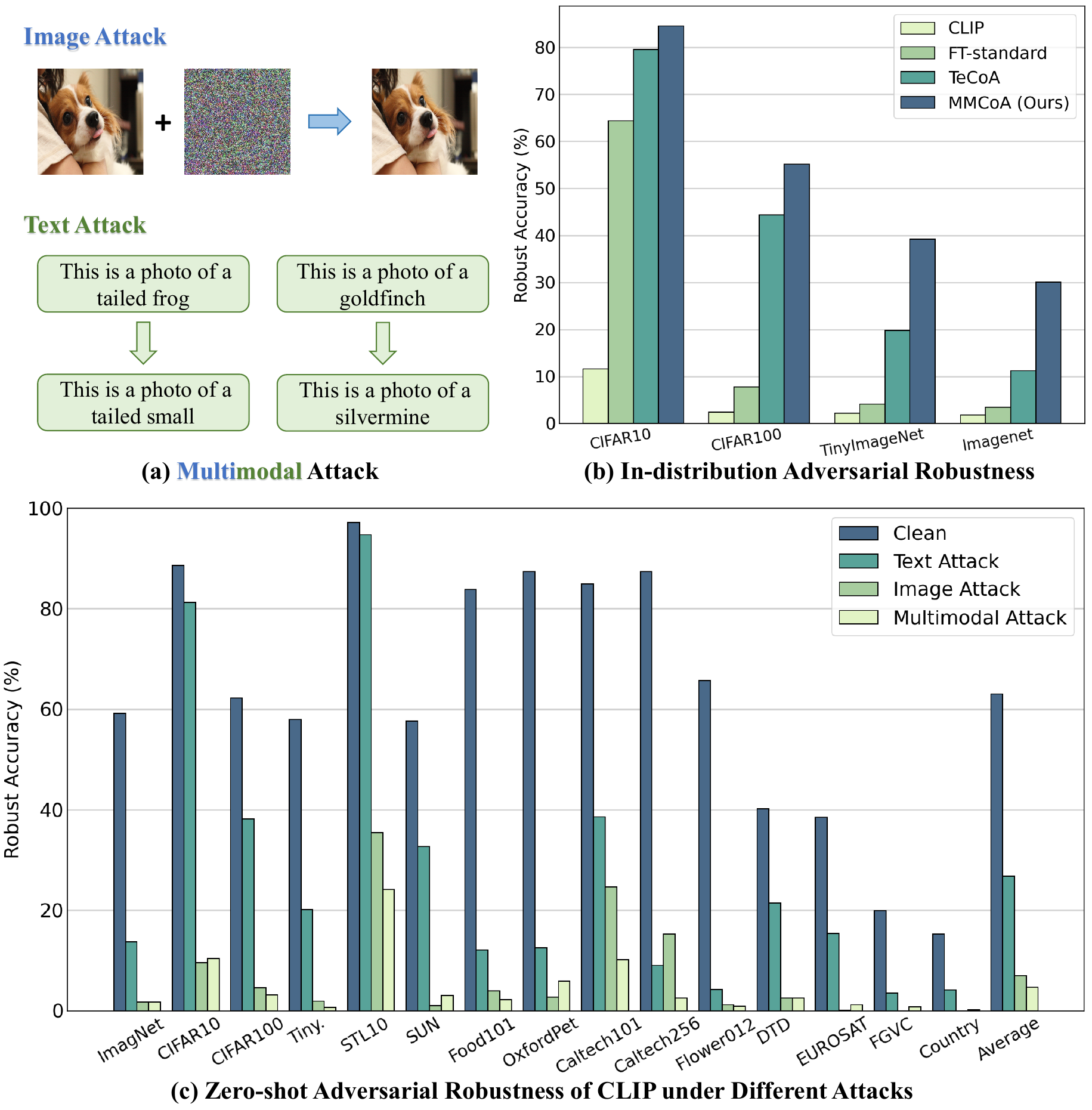}
  \caption{Adversarial attacks and adversarial robustness. (a) An example of the multimodal adversarial attack. (b) Accuracies of different methods for in-distribution adversarial robustness under the multimodal attack. (c) Accuracies of CLIP for zero-shot adversarial robustness under different attacks.}
  \label{fig:intro}
\end{figure}

\IEEEPARstart{P}{\lowercase{retrained}} vision-language models~(VLMs) have demonstrated remarkable generalization capabilities across a wide range of downstream tasks, and have become pivotal tools in the emerging field of multimodal artificial intelligence, including image classification~\cite{radford2021learning,jia2021scaling}, semantic segmentation~\cite{luddecke2022image} and robotics~\cite{zitkovich2023rt}. 
Given that a growing number of vision-language models are utilized in security-sensitive downstream tasks, enhancing the robustness of such models has become an urgent priority.

In this sense, the robustness of the CLIP model \cite{radford2021learning}, a foundational framework for many VLMs, has become a growing area of research. 
While existing work has primarily addressed adversarial defenses against image-based attacks in image classification tasks \cite{zhang2023adversarial,mao2022understanding,wang2024pre,zhao2024evaluating}, a significant gap remains when it comes to the understanding and mitigating vulnerabilities to textual and multimodal adversarial attacks. 
Methods such as Bert-Attack \cite{li2020bert}, Co-Attack \cite{zhang2022towards}, and SGA \cite{lu2023setlevel} highlight this vulnerability, underscoring the need for more comprehensive research.
\textit{Building upon these insights, our work presents the first comprehensive study aimed at improving the adversarial robustness of VLMs against attacks targeting different modalities: image, text, and multi-modality.}
We investigate supervised classification tasks to address these vulnerabilities. 
While our primary focus is on classification tasks, we also conduct a broader evaluation of adversarial robustness in tasks such as image-text retrieval, particularly under multimodal attacks like Co-Attack, as demonstrated in our experiments.

In this study, we first combine an image-based PGD attack~\cite{madry2017towards} with a text-based BERT-Attack~\cite{li2020bert} to generate multimodal adversarial examples against the CLIP model, as shown in Fig.~\ref{fig:intro} (a).
Next, we apply these attacks to the CLIP model, and our observations, depicted in Fig.~\ref{fig:intro} (c), reveal that:
(1) imperceptible perturbations within the image domain significantly reduce the model's performance;
(2) Under BERT-Attack, the more categories a dataset contains, the stronger the attack's efficacy. 
For instance, the accuracy decrease in CIFAR10 is less pronounced compared to the more substantial decline observed in CIFAR100.
Consequently, enhancing the adversarial robustness of text encoders in VLMs emerges as a critical challenge.
Therefore, an important question arises: \textit{how can we effectively address the challenges posed by multimodal adversarial examples to improve the robustness of VLMs?}
 
Adversarial training~\cite{chen2020adversarial} is regarded as one of the most effective defense strategies. 
Previous methods mainly adopt adversarial training into prompt tuning and fine-tuning to defend against image-based attacks~\cite{zhang2023adversarial, mao2022understanding}. 
However, as shown in Fig.~\ref{fig:intro} (b), these methods prove insufficient when confronted with multimodal attacks.
This is because existing methods primarily focus on defending CLIP against adversarial images, neglecting the synergistic effect of combined image and text attacks. 
A successful multimodal attack might leverage subtle perturbations in both modalities, resulting in a misalignment that single-modality defenses cannot effectively counteract.
To address this challenge and enhance the adversarial robustness of CLIP, we propose a simple yet effective \textbf{M}ulti\textbf{m}odal \textbf{Co}ntrastive \textbf{A}dversarial (MMCoA) training framework.
The framework introduces two components: a text-supervised image adversarial loss and an image-supervised text adversarial loss. 
These two losses ensure that clean text embeddings remain aligned with adversarial image embeddings, while clean image embeddings stay aligned with adversarial text embeddings. 
Such an approach creates a robust visual-text joint space by maintaining cross-modal alignment under adversarial attacks, making the model more resilient to distortions and preserving its ability to associate visual and textual data.
As illustrated in Fig.~\ref{fig:intro} (b), our MMCoA markedly enhances the multimodal adversarial robustness of CLIP.

In summary, we conduct an in-depth investigation of three types of attacks on 15 datasets, offering a comprehensive study assessing their impact on adversarial robustness across two distinct tasks. 
The first task is to explore in-distribution adversarial robustness, which tests defensive performance within the same dataset. 
The second task is to explore out-of-distribution generalization adversarial robustness, also referred to as zero-shot adversarial robustness, which evaluates defensive performance for unknown tasks.
Our extensive experiments and analysis reveal a number of novel and intriguing insights, paving the way for a unified framework in adversarial robustness against different modality attacks, and opening up new possibilities for securing VLMs against multimodal attacks.

Our main contributions can be summarized as follows:

\begin{itemize}
    \item To the best of our knowledge, we conduct the first comprehensive study aimed at enhancing the adversarial robustness of VLMs against attacks targeting visual, textual, and multimodal inputs.
    \item We propose the \textbf{M}ulti\textbf{m}odal \textbf{Co}ntrastive \textbf{A}dversarial training (MMCoA), which effectively enhances the adversarial robustness of both image and text encoders.
    \item Extensive experiments and detailed analysis on 15 datasets for two tasks under three types of attacks reveal the performance and characteristics of different adversarial defense methods, providing valuable insights for enhancing the security of VLMs.
\end{itemize}

\section{Related Work}
\label{sec:rw}

\subsection{Adversarial Attacks on Image, Text, and Multimodal Data}
Adversarial attacks were first introduced in computer vision, where perturbed images were crafted to deceive models into making incorrect classifications~\cite{szegedy2013intriguing, 10385167}. 
Gradient-based adversarial attacks have been extensively studied, including methods such as FGSM~\cite{goodfellow2014explaining}, BIM and ILCM~\cite{kurakin2018adversarial}, and PGD~\cite{madry2017towards}. 
The key idea behind gradient-based methods is to find minimal perturbations that maximize the likelihood of incorrect predictions, achieved by applying gradient descent over the continuous space of images.
In natural language processing, successful adversarial attacks typically use heuristic rules to modify characters in a word~\cite{jin2019bert, li2020bert} or substitute words with synonyms~\cite{li2018textbugger, ren2019generating}. 

The advent of VLMs introduces a new dimension: attacks can target either the image input, the text input, or both simultaneously. 
This multimodal attack landscape significantly expands the potential vulnerabilities compared to attacks on images or text alone. 
It is important to clarify that in this work, multimodal adversarial examples refer to instances where both the image and text modalities are attacked.
Current methods for generating multimodal adversarial examples can be categorized into two approaches. 
The first involves attacking each modality independently, with no interaction between adversarial modalities, as in Sep-Attack~\cite{lu2023setlevel}. 
In the second approach, adversarial examples from different modalities interact, where by perturbations in one modality influence those in the other. 
Examples include Co-Attack~\cite{zhang2022towards} and SGA~\cite{lu2023setlevel}, which perturb image-text pairs simultaneously to maximize the Kullback-Leibler (KL) divergence between the clean and adversarial multimodal embeddings.
Specifically, Co-Attack generates adversarial images by maximizing the KL divergence between the clean and adversarial image embeddings, as well as between the adversarial image and adversarial text embeddings.
These attacks pose a considerable threat to the robustness and reliability of VLMs, as they can compromise model performance across various tasks.
Driven by the need for defense, we develop a novel framework to tackle adversarial attacks across visual, textual, and multimodal inputs.

\begin{figure*}[ht]
  \centering
  \includegraphics[width=0.85\textwidth]{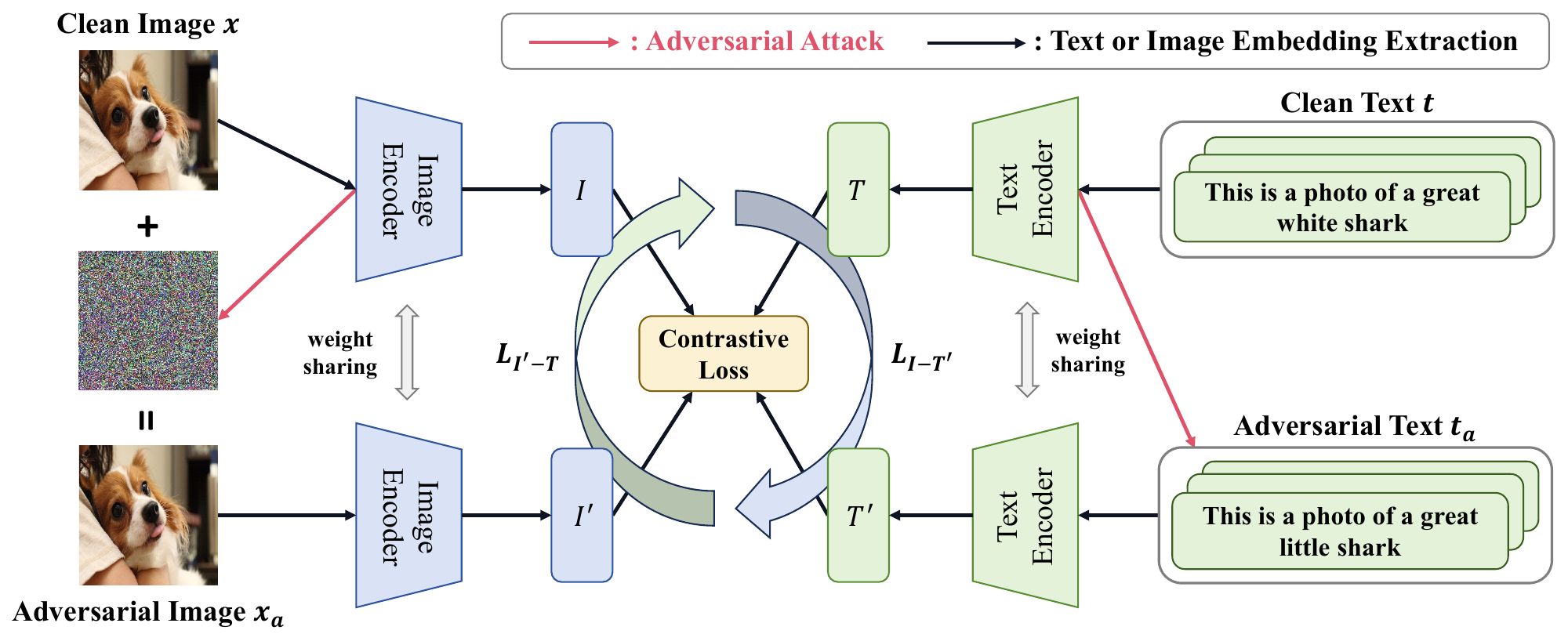}
  \caption{Overview of our proposed Multimodal Contrastive Adversarial (MMCoA) training framework. To achieve multimodal adversarial robustness, we extend the adversarial training paradigm to the joint training of adversarial examples for both images and texts by adversarial contrastive learning with vision and language supervision.}
  \label{fig:model}
\end{figure*}

\subsection{Adversarial Training on Visual-Language Models}
Deep neural networks have been shown to be vulnerable to adversarial attacks, prompting the development of various defense strategies, such as adversarial training~\cite{moosavi2016deepfool, sankaranarayanan2018regularizing, jin2023randomized, 10124732,lee2023advancing,10292990}, data transformation~\cite{dziugaite2016study, guo2017countering, bhagoji2018enhancing}, and network modifications~\cite{gu2014towards, ross2018improving}. 
Among these, adversarial training has emerged as one of the most effective techniques for enhancing the adversarial robustness of models by incorporating adversarial examples directly into the training process.

Recent work has extended adversarial training to VLMs~\cite{zhang2023adversarial, mao2022understanding}. However, these studies primarily focus on image attacks, where adversarial examples are generated from images and used to fine-tune the models. 
AdvPT~\cite{zhang2023adversarial} differs slightly from traditional adversarial training in that its adversarial examples are generated prior to the training phase. 
AdvPT first generates adversarial images by maximizing the KL divergence between the clean and adversarial image embeddings, which are specifically obtained from CLIP's image encoder. 
During the training phase, it optimizes learnable textual prompts to align with the adversarial image embeddings.
TeCoA~\cite{mao2022understanding} introduces a text-guided adversarial training method, where adversarial images are generated by maximizing the contrastive loss between adversarial image embeddings and clean text embeddings, and the image encoder of CLIP is updated by optimizing this loss.
Although TeCoA~\cite{mao2022understanding} does not explicitly address defenses against multimodal adversarial examples, it remains an important baseline in our work.
This approach has been shown to outperform other adversarial training strategies, such as standard adversarial training, which adds a learnable linear layer to CLIP's image encoder and generates adversarial images by maximizing the classification loss on clean images and their corresponding one-hot labels. 
It also surpasses non-text-guided contrastive adversarial training, where adversarial images are generated by maximizing the contrastive loss between the clean and adversarial image embeddings or between adversarial image embeddings and the one-hot labels.

In contrast to these methods, our study addresses the more complex challenge of defending against multimodal adversarial attacks, targeting both the image and text encoders in VLMs like CLIP.
This is achieved by proposing a multimodal contrastive adversarial training (MMCoA) framework, which aligns clean text embeddings with adversarial image embeddings, and adversarial text embeddings with clean image embeddings, thereby ensuring consistency across modalities even in the presence of adversarial perturbations.
Moreover, we evaluate both in-distribution and zero-shot robustness under multimodal adversarial training.
\textit{Consequently, we present the first comprehensive study that aims at improving adversarial robustness against attacks targeting different modalities: image, text, and image-text multi-modality.}

\subsection{Adapting Vision Language Models}
Adapting the vision-language model CLIP for specific downstream tasks typically involves a series of fine-tuning approaches. 
For instance, linear probing~\cite{radford2021learning} adds a classifier on top of CLIP's image encoder. 
Partial fine-tuning updates the last few layers of the model, while full fine-tuning updates the entire model. 
Recently, parameter-efficient fine-tuning methods have been extensively studied~\cite{zhou2022learning, jia2022visual, khattak2023maple}. 
These methods introduce only a small set of parameters while keeping the main model fixed, such as prompt tuning.
Prompt tuning involves adding tunable parameters to the model's input, which are optimized in a data-driven fashion via backpropagation, and has been extensively explored in many downstream tasks~\cite{shu2022test, cho2023promptstyler, bai2024prompt}.
However, in the context of adversarial learning, the adaptation or defense of large-scale VLMs against attacks on multimodal and text data remains rather unexplored.
To this end, we aim to verify the effectiveness of multimodal adversarial training, which can be extended to all tuning methods. 
For fair comparisons, our primary methodological framework and all baselines utilize full fine-tuning.
Additionally, we also compare adversarial prompt tuning methods in the Appendix.
Our findings suggest that our MMCoA method provides stronger robustness compared to unimodal adversarial fine-tuning, potentially opening new avenues for enhancing the security of VLMs.

\begin{table*}[htbp]
\centering 
\caption{In-distribution robust accuracies across 4 datasets under \textbf{three types of attacks}, \textit{i.e.,} image attack, text attack and multimodal attack. Except for CLIP, we fine-tune all methods on each dataset and then test them on the same dataset. For image attacks, we utilize 100 steps of PGD, while text attacks are conducted using BERT-Attack. $\Delta$ denotes the difference in accuracy between our method and a baseline. Bold font denotes the best accuracies.}
\label{tab:iid}
\begin{tabular}{l|cccc|cccc}
\midrule
\multicolumn{1}{l|}{\multirow{2}{*}{Method}}  & \multicolumn{4}{c|}{CIFAR10}  & \multicolumn{4}{c}{CIFAR100} \\
\cmidrule{2-9} 
& Image Attack & Text Attack & Multimodal Attack & Clean & Image Attack & Text Attack & Multimodal Attack & Clean \\
\midrule
CLIP & 9.32 & 79.74 & 11.65 & 88.56 & 4.59 & 37.24 & 2.42 & 62.28 \\
FT-standard & 63.93 & \textbf{96.21} & 64.39 & \textbf{97.01} & 6.53 & 62.28 & 7.74 & \textbf{85.21} \\
TeCoA & \textbf{85.06} & 92.04 & 79.53 & 95.88 & \textbf{60.71} & 56.96 & 44.34 & 80.01 \\
MMCoA & 84.55 & 96.16 & \textbf{84.52} & 96.17 & 59.50 & \textbf{75.38} & \textbf{55.13} & 81.51 \\
\midrule
$\Delta(\text{CLIP})$ & \textcolor[cmyk]{1, 0, 1, 0.61}{+75.23} & \textcolor[cmyk]{1, 0, 1, 0.61}{+16.42} & \textcolor[cmyk]{1, 0, 1, 0.61}{+72.87} & \textcolor[cmyk]{1, 0, 1, 0.61}{+7.61} & \textcolor[cmyk]{1, 0, 1, 0.61}{+54.91} & \textcolor[cmyk]{1, 0, 1, 0.61}{+38.14} & \textcolor[cmyk]{1, 0, 1, 0.61}{+52.71} & \textcolor[cmyk]{1, 0, 1, 0.61}{+19.23} \\
$\Delta(\text{SOTA})$ & \textcolor[cmyk]{0, 0.75, 0.75, 0.35}{-0.51} & \textcolor[cmyk]{0, 0.75, 0.75, 0.35}{-0.05} & \textcolor[cmyk]{1, 0, 1, 0.61}{+4.99} & \textcolor[cmyk]{0, 0.75, 0.75, 0.35}{-0.84} & \textcolor[cmyk]{0, 0.75, 0.75, 0.35}{-1.21} & \textcolor[cmyk]{1, 0, 1, 0.61}{+13.10} & \textcolor[cmyk]{1, 0, 1, 0.61}{+10.79} & \textcolor[cmyk]{0, 0.75, 0.75, 0.35}{-3.70}  \\
\midrule
\multicolumn{1}{l|}{\multirow{2}{*}{Method}}  & \multicolumn{4}{c|}{TinyImagenet}  & \multicolumn{4}{c}{ImageNet} \\
\cmidrule{2-9} 
& Image Attack & Text Attack & Multimodal Attack & Clean & Image Attack & Text Attack & Multimodal Attack & Clean \\
\midrule
CLIP & 1.93 & 19.67 & 2.18 & 59.46 & 0.79 & 13.65 & 1.80 & 59.16 \\
FT-standard & 4.55 & 29.96 & 4.10 & \textbf{78.96} & 3.88 & 17.61 & 3.47 & 64.57 \\
TeCoA & 49.01 & 25.09 & 19.80 & 66.25 & 41.47 & 15.84 & 11.96 & 63.29 \\
MMCoA & \textbf{53.97} & \textbf{54.13} & \textbf{39.19} & 76.11 & \textbf{46.58} & \textbf{41.34} & \textbf{30.02} & \textbf{67.79} \\
\midrule
$\Delta(\text{CLIP})$ & \textcolor[cmyk]{1, 0, 1, 0.61}{+52.04} & \textcolor[cmyk]{1, 0, 1, 0.61}{+34.46} & \textcolor[cmyk]{1, 0, 1, 0.61}{+37.01} & \textcolor[cmyk]{1, 0, 1, 0.61}{+16.65} & \textcolor[cmyk]{1, 0, 1, 0.61}{+45.79} & \textcolor[cmyk]{1, 0, 1, 0.61}{+27.69} & \textcolor[cmyk]{1, 0, 1, 0.61}{+28.22} & \textcolor[cmyk]{1, 0, 1, 0.61}{+8.63} \\
$\Delta(\text{SOTA})$ & \textcolor[cmyk]{1, 0, 1, 0.61}{+4.96} & \textcolor[cmyk]{1, 0, 1, 0.61}{+24.17} & \textcolor[cmyk]{1, 0, 1, 0.61}{+19.39} & \textcolor[cmyk]{0, 0.75, 0.75, 0.35}{-2.85} & \textcolor[cmyk]{1, 0, 1, 0.61}{+5.11}  & \textcolor[cmyk]{1, 0, 1, 0.61}{+25.50} & \textcolor[cmyk]{1, 0, 1, 0.61}{+18.06} & \textcolor[cmyk]{1, 0, 1, 0.61}{+3.22} \\
\bottomrule
\end{tabular}
\end{table*}

\section{Multimodal Defense for VLMs}
\label{sec:method}

\subsection{Background and Problem Setup}
\label{subsec:BPS}

\noindent \textbf{Revisiting CLIP.} 
Our study of multimodal adversarial robustness builds upon a pre-trained vision-language model CLIP~\cite{radford2021learning}.
Contrastive Language-Image Pre-Training (CLIP) model is pre-trained on 400 million image-text pairs collected from the internet with contrastive learning.
It is composed of an image encoder $f$ and a text encoder $g$, which are adopted to encode images $x$ and their corresponding natural language descriptions $t$, respectively.

For adapting CLIP to downstream tasks, the natural language descriptions $t$ are usually manually designed as ``This is a photo of a [CLS]", where [CLS] denotes the class labels from a dataset.
For zero-shot classification, the prediction $\hat{y}$ of an image $x$ is calculated as

\begin{equation}\label{func:zs}
p(\hat{y} \mid x)=\frac
{\exp (\langle g(t_{\hat{y}}), f(x) \rangle / \tau)}
{\sum^K_{j=1} \exp (\langle g(t_j), f(x) \rangle / \tau)},
\end{equation}
where $\langle\cdot,\cdot\rangle$ denotes the cosine similarity, $\tau$ designates the temperature parameter used to control the sharpness of the distribution, $K$ represents the number of classes, $g(t_{\hat{y}})$ refers to the text embedding while $f(x)$ represents the image embedding. 

\noindent \textbf{Multimodal Attack.} 
In this study, the multimodal attack is combined with the image-based PGD attack~\cite{madry2017towards} and text-based BERT-Attack~\cite{li2020bert}. 

The Projected Gradient Descent (PGD) attack is a popular adversarial attack method designed to create adversarial examples that deceive machine learning models. 
It iteratively applies gradient ascent techniques on the input data to maximize a chosen loss function $\mathcal{L}$ (e.g., cross-entropy loss), thereby generating an adversarial sample within a predefined perturbation limit that aims to maximize the error rate of the model.
The core of the PGD algorithm unfolds through iterative steps, which can be formulated as

\begin{equation}\label{func:image-attack}
x_{i+1}=\Pi_{x+S}\left(x_i+\alpha \cdot \operatorname{sign}\left(\nabla_x \mathcal{L}\left(\theta, x_i, y\right)\right)\right),
\end{equation}
where \( \theta \) denotes the model parameters, $y$ represents the ground-truth label, \( \alpha \) is the step size, and \( \Pi_{x+S}\) refers to the projection onto the perturbation set defined by the \( \epsilon \)-ball around \( x \) under the \( \mathcal{L}_{\infty} \) or \( \mathcal{L}_2 \) norm, ensuring the adversarial perturbation remains within the acceptable limits.
The PGD attack method seeks to find an optimal yet human-imperceptible perturbation that misleads neural models with predefined bounds.
In this study, we apply the PGD method to generate adversarial images targeting the CLIP model. Specifically, we modify the loss function in Equation~\eqref{func:image-attack} to utilize CLIP's contrastive loss, which can be formulated as

\begin{equation}\label{func:image-attack-CLIP}
x_{i+1}=\Pi_{x+S}\left(x_i+\alpha \cdot \operatorname{sign}\left(\nabla_x \mathcal{L}_c\left(\theta, x_i, t\right)\right)\right),
\end{equation}
where $\mathcal{L}_c$ is same as the contrastive loss in Equation \eqref{func:image_loss}.
It is computed between the clean text embedding and the adversarial image embedding, with the text serving as the supervised target.

The BERT-Attack \cite{li2020bert} is an adversarial attack method tailored for the BERT model or other models with transformer-based architectures.
BERT-Attack targets text data, generating adversarial examples by modifying, inserting, or deleting words in the input text $t$. 
The core idea of BERT-Attack leverages a pre-trained language model to identify and replace vulnerable words in the input text $t$ with semantically consistent alternatives that have the most significant impact on the target model's prediction, thereby maximizing the target model's output error.
The adversarial text input is obtained by maximizing the divergence in the feature space between the perturbed text and the original text, employing measures such as KL divergence, that is 
\begin{equation}\label{func:text-attack}
\begin{aligned}
&t_a = R(t), \\
&\text{s.t.}~ t_a = \underset{t_a}{\arg \max }\left(\left\|g\left(t_a\right)-g\left(t\right)\right\|\right),
\end{aligned} 
\end{equation}
where $R (\cdot)$ denotes the operation of replacing or modifying tokens in the input text, and $g (\cdot)$ designates the target text encoder. 
As shown in Fig.~\ref{fig:model}, given that class tokens have a more pronounced impact on the model's prediction, replacements predominantly occur with these tokens.

\begin{table*}[htbp]
\centering
\caption{Out-of-distribution robust accuracies across 15 datasets under \textbf{multimodal attack}. Except for CLIP, we fine-tune all methods on the ImageNet dataset with the few-shot setting (1-shot, 5-shot, and 50-shot) and full-shot setting, and then test them on the remaining datasets. For image attacks, we utilize 100 steps of PGD, while text attacks are conducted using BERT-Attack. Bold font denotes the best accuracies.}
\label{tab:ood}
\resizebox{\textwidth}{!}{
\begin{tabular}{lcccccccccccccccc}
\toprule
  & Source & \multicolumn{15}{c}{Target} \\
\cmidrule{2-17} 
Method & \rotatebox{90}{ImagNet} & \rotatebox{90}{CIFAR10} & \rotatebox{90}{CIFAR100} & \rotatebox{90}{Tiny.} & \rotatebox{90}{STL10} & \rotatebox{90}{SUN397} & \rotatebox{90}{Food101} & \rotatebox{90}{OxfordPets} & \rotatebox{90}{Flower102} & \rotatebox{90}{DTD} & \rotatebox{90}{EuroSAT} & \rotatebox{90}{FGVCA.} & \rotatebox{90}{Country211} & \rotatebox{90}{Caltech101} & \rotatebox{90}{Caltech256} & \rotatebox{90}{Average} \\
\midrule
CLIP & 1.75 & 10.36 & 3.13 & 0.69 & 24.08 & 3.07 & 2.21 & 5.89 & 0.92 & 2.55 & 1.20 & 0.84 & 0.21 & 10.16 & 2.55 & 4.64 \\
\midrule

\textit{1-shot} &  &  &  &  &  &  &  &  &  &  &  &  &  &  &  &  \\
\midrule
FT-standard & 1.19 & 8.93 & \textbf{4.82} & 0.50 & 24.34 & 3.43 & \textbf{2.70} & 4.69 & 1.18 & 2.18 & 0.90 & 1.11 & \textbf{0.27} & \textbf{11.93} & \textbf{2.62} & 4.72 \\
TeCoA & 1.77 & 10.05 & 3.18 & \textbf{0.73} & 23.84 & 3.06 & 2.18 & 5.78 & 0.99 & 2.34 & \textbf{1.02} & 0.84 & 0.26 & 10.26 & 2.56 & 4.59 \\
MMCoA & \textbf{1.83} & \textbf{10.33} & 3.22 & 0.64 & \textbf{25.24} & \textbf{3.45} & 2.19 & \textbf{6.00} & \textbf{1.38} & \textbf{3.14} & 0.81 & 0.84 & \textbf{0.27} & 11.12 & \textbf{2.62} & \textbf{4.87} \\
\midrule

\textit{5-shot} &  &  &  &  &  &  &  &  &  &  &  &  &  &  &  &  \\
\midrule
FT-standard & 1.58 & 9.13 & 4.67 & 0.41 & 26.72 & 2.66 & 2.00 & \textbf{5.97} & 0.93 & 2.93 & 0.45 & 0.69 & 0.18 & 11.63 & 2.72 & 4.84 \\
TeCoA & 2.55 & 17.69 & 4.42 & 0.39 & 38.39 & 4.52 & 1.45 & 4.14 & 1.63 & 2.93 & 0.80 & 0.96 & 0.36 & 13.94 & 2.24 & 6.43 \\
MMCoA & \textbf{4.47} & \textbf{24.44} & \textbf{11.44} & \textbf{1.08} & \textbf{48.18} & \textbf{8.89} & \textbf{2.34} & 5.21 & \textbf{2.98} & \textbf{6.76} & \textbf{4.66} & \textbf{1.56} & \textbf{0.66} & \textbf{20.81} & \textbf{4.06} & \textbf{9.84} \\
\midrule

\textit{50-shot} &  &  &  &  &  &  &  &  &  &  &  &  &  &  &  &  \\
\midrule
FT-standard & 3.12 & 16.65 & 7.99 & 0.45 & 39.09 & 4.81 & 1.85 & 6.00 & 1.07 & 4.14 & 0.25 & 0.84 & 0.35 & 13.50 & 3.23 & 6.89 \\
TeCoA & 7.17 & 40.97 & 16.28 & 5.29 & 68.94 & 17.64 & 4.15 & \textbf{6.49} & 3.45 & 11.86 & 1.87 & 2.07 & 1.12 & 31.80 & 5.93 & 15.00 \\
MMCoA & \textbf{10.47} & \textbf{57.44} & \textbf{22.41} & \textbf{11.46} & \textbf{76.74} & \textbf{18.70} & \textbf{4.93 }& 4.09 & \textbf{3.58} & \textbf{16.33} & \textbf{4.91} & \textbf{2.46} & \textbf{1.41} & \textbf{35.63} & \textbf{6.34} & \textbf{18.46} \\
\midrule

\textit{Full-shot} &  &  &  &  &  &  &  &  &  &  &  &  &  &  &  &  \\
\midrule
FT-standard & 3.47 & 16.77 & 8.11 & 1.24 & 41.16 & 4.95 & 1.60 & 5.61 & 1.45 & 3.24 & 0.32 & 1.47 & 0.36 & 15.73 & 2.95 & 7.23 \\
TeCoA & 11.96 & 54.40 & 25.54 & 12.43 & 78.18 & 21.62 & \textbf{7.12} & 12.40 & \textbf{2.57} & \textbf{15.05} & 10.56 & \textbf{2.55} & 1.43 & 36.49 & 7.15 & 19.96 \\
MMCoA & \textbf{22.89} & \textbf{61.47} & \textbf{29.31} & \textbf{17.66} & \textbf{82.99} & \textbf{21.94} & 4.53 & \textbf{15.59} & 2.49 & 7.71 & \textbf{14.07} & 1.86 & \textbf{1.53} & \textbf{38.69} & \textbf{7.23} & \textbf{22.00} \\
\bottomrule
\end{tabular}
}
\end{table*}

\noindent \textbf{Adversarial Training} is a typical method for training deep neural networks that are robust to adversarial attacks.
It is formulated as a min-max optimization problem, wherein adversarial examples are crafted during the maximization step (previously defined). 
The subsequent minimization step then optimizes the model parameters to reduce the loss on these adversarial examples. 
Formally, this minimization step can be expressed as

\begin{equation}\label{func:adv}
\theta = \arg\min _\theta \mathcal{L}(\theta, x_a, y).
\end{equation}

Previous methods have mainly concentrated on defending against image-based attacks by fine-tuning models with adversarially generated image examples.
For large-scale vision-language models, it is essential not only to defend against image attacks but also to counteract textual attacks. 
Therefore, the minimization process of the adversarial training paradigm extends to the joint training of adversarial examples for both images and texts, which can be formulated as

\begin{equation}\label{func:adv_vlm}
\theta = \arg\min _\theta \mathcal{L}(\theta, x_a, t_a, y).
\end{equation}

\subsection{Multimodal Contrastive Adversarial Training}
To improve adversarial robustness in both image and text encoders, two strategies can be considered: aligning adversarial text embeddings with clean text embeddings and adversarial image embeddings with clean image embeddings, or aligning adversarial text embeddings with adversarial image embeddings. However, the first approach lacks cross-modal alignment, while the second aligns adversarial examples from both modalities, which offers no useful guidance.
Instead, we propose a simple yet effective \textbf{M}ulti\textbf{m}odal \textbf{Co}ntrastive \textbf{A}dversarial (MMCoA) training method, as shown in Fig.~\ref{fig:model}. 
The MMCoA is outlined as follows.

\noindent \textbf{Text-supervised Image Adversarial Training.}
For each image, we utilize the images and the corresponding manually crafted prompts, ``This is a photo of a [CLS]" to form image-text pairs.
We first generate adversarial images using an image-based attack (Section~\ref{subsec:BPS}). 
As shown in Fig.~\ref{fig:intro} (c), this misalignment degrades the model's ability to associate manipulated visual data with correct text. 
Text-supervised image adversarial training aims to minimize the distance between the feature embeddings of the attacked image $x_a$ and the correct corresponding text inputs $t$.
Subsequently, we apply the text-supervised image adversarial contrastive loss, which can formulated as

\begin{equation}
\begin{aligned}\label{func:image_loss}
&\mathcal{L}_{I'-T}({x_a}, t, \mathbf{y})=-\mathbb{E}_{i, j}\left[\mathbf{y}_{i j} \log \frac{\exp \left(\langle I'_i, T_j \rangle / \tau\right)}{\sum_k \exp \left(\langle I'_i, T_k \rangle / \tau\right)}\right], \\
\end{aligned} 
\end{equation}
where $I'=f(x_a)$ denotes the embeddings of the adversarial input images, $T=g(t)$ are clean text embeddings, and $\mathbf{y}$ designates the contrastive labels for the image-text pairs, $\mathbf{y}_{ij} = 1$ if $i = j$ and $\mathbf{y}_{ij} = 0$ otherwise. 
In this way, we align the adversarial image embeddings with their corresponding clean text embeddings.

\noindent \textbf{Image-supervised Text Adversarial Training.} Similar to text-supervised image adversarial training, we first transform the manually crafted prompt into adversarial text examples. 
To defend the text attacks, the image-supervised text adversarial training minimizes the distance between the feature embeddings of the clean image $x$ and the adversarial corresponding text inputs $t_a$. 
Then, we apply the image-supervised text adversarial contrastive loss as

\begin{equation}
\begin{aligned}\label{func:text_loss}
&\mathcal{L}_{I-T'}({x}, t_a, \mathbf{y})=-\mathbb{E}_{i, j}\left[\mathbf{y}_{i j} \log \frac{\exp \left(\langle I_i, T'_j \rangle / \tau\right)}{\sum_k \exp \left(\langle I_i, T'_k \rangle / \tau\right)}\right], \\
\end{aligned}
\end{equation}
where $I=f(x)$ denotes the embeddings of the clean input images, and $T'=g(t_a)$ designates adversarial text embeddings, with $\mathbf{y}_{ij} = 1$ if $i = j$ and $\mathbf{y}_{ij} = 0$ otherwise.
In this way, the adversarial text embeddings are aligned with their corresponding clean image embeddings.

\begin{figure*}[htbp]
  \centering
  \includegraphics[width=0.95\textwidth]{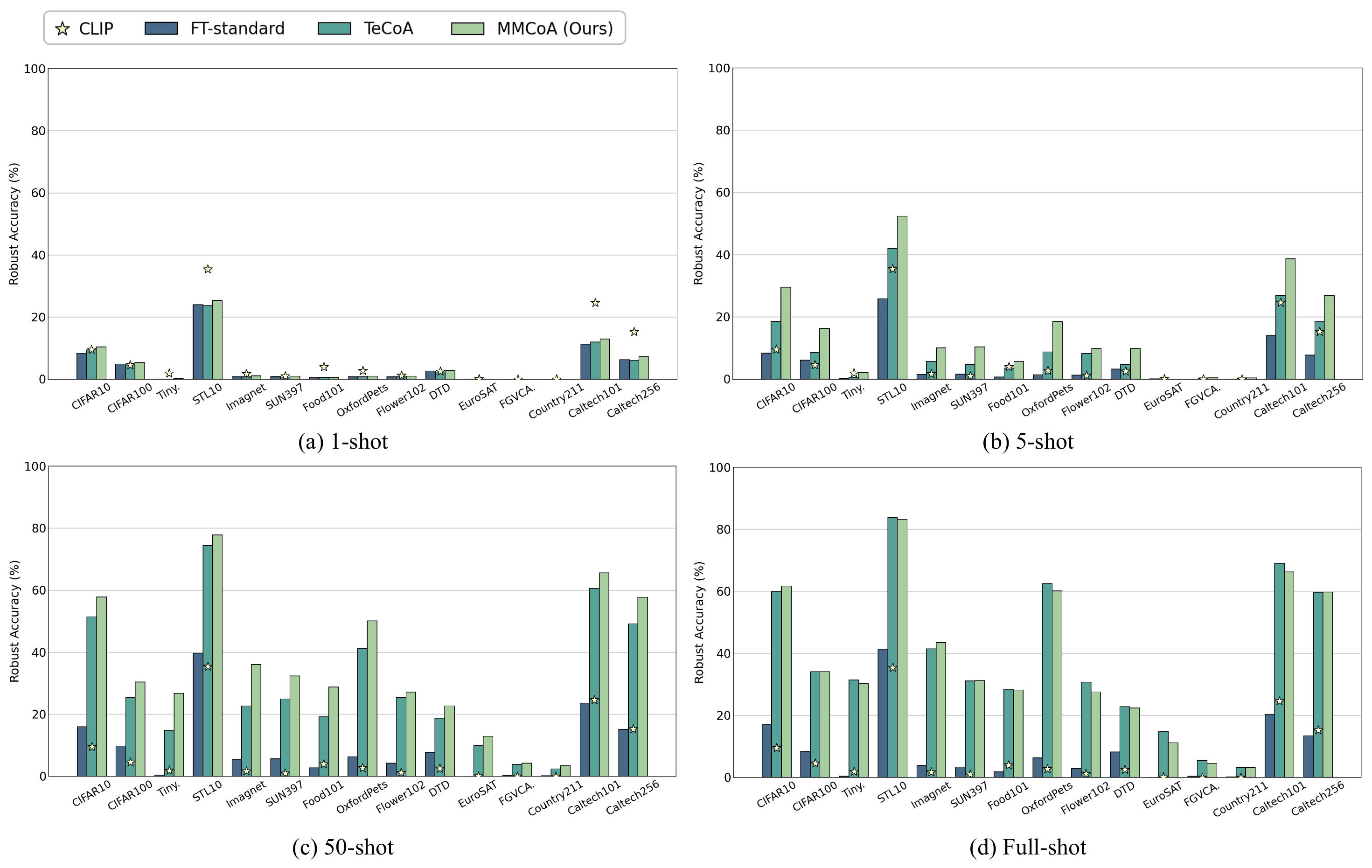}
  \caption{Out-of-distribution robust accuracies across 15 datasets under 100 steps of the PGD \textbf{image attack}. We fine-tuned all methods on the ImageNet dataset with the few-shot setting and full-shot setting, and then tested them on the remaining datasets.}
  \label{fig:ood_image}
\end{figure*}

\begin{algorithm}
\caption{MMCoA Training for Adversarial Defense in Multi-Modal Models}
\label{algo:MMCoA}
\begin{algorithmic}[1]
\REQUIRE CLIP model $M$, images $x$ and class names $c$, perturbation bound $\epsilon$, step size $\alpha$, number of steps $p$, learnable parameter $\theta$, pre-trained Bert Model $B$
\STATE Initialize class text prompts $t$ for each class [CLS] in $c$ as $t.\text{append}('\text{This is a photo of a }' + [\text{CLS}])$
\STATE Initialize $\theta$ with the pre-trained parameters from CLIP $M$
\FOR{each iteration in training epochs}
\FOR{each $(x,t)$ in minibatch}
\STATE $x_{ta} = \text{ImageAttack}(x, t, \epsilon, \alpha, p, \theta)$ \COMMENT{Generate adversarial image samples using the method depicted in Section~\ref{subsec:BPS}}
\STATE $t_{a} = \text{BertAttack}(t, B, \theta)$ \COMMENT{Generate adversarial text samples using Equation \eqref{func:text-attack}}
\STATE Optimize $\theta$ to minimize $\mathcal{L}$ as defined in Equation \eqref{func:total_loss}
\ENDFOR
\ENDFOR
\STATE \textbf{return} $\theta$ \COMMENT{Return the optimized parameters}
\end{algorithmic}
\end{algorithm}

As a result, our MMCoA method can be trained end-to-end with a total contrastive loss, which can be formulated as:

\begin{equation}\label{func:total_loss}
\mathcal{L} = \gamma_1\mathcal{L}_{I'-T} + \gamma_2\mathcal{L}_{I-T'},
\end{equation}
where $\gamma_1$ and $\gamma_2$ are hyper-parameters.
Our method iteratively alternates between generating adversarial examples and updating the model, with the updates guided by Equation~\eqref{func:adv_vlm} and Equation~\eqref{func:total_loss}. 

The overall process of MMCoA is summarized in Algorithm~\ref{algo:MMCoA}. 
The adversarial robustness of both the image encoder and the text encoder is significantly enhanced through the strategic alignment of clean text embeddings with the image embeddings of adversarial images and the alignment of adversarial text embeddings with the image embeddings of clean images. 
This dual alignment fosters a robust multimodal representation, strengthening the model's defenses against adversarial attacks targeting different modalities.

\begin{figure*}[htbp]
  \centering
  \includegraphics[width=0.95\textwidth]{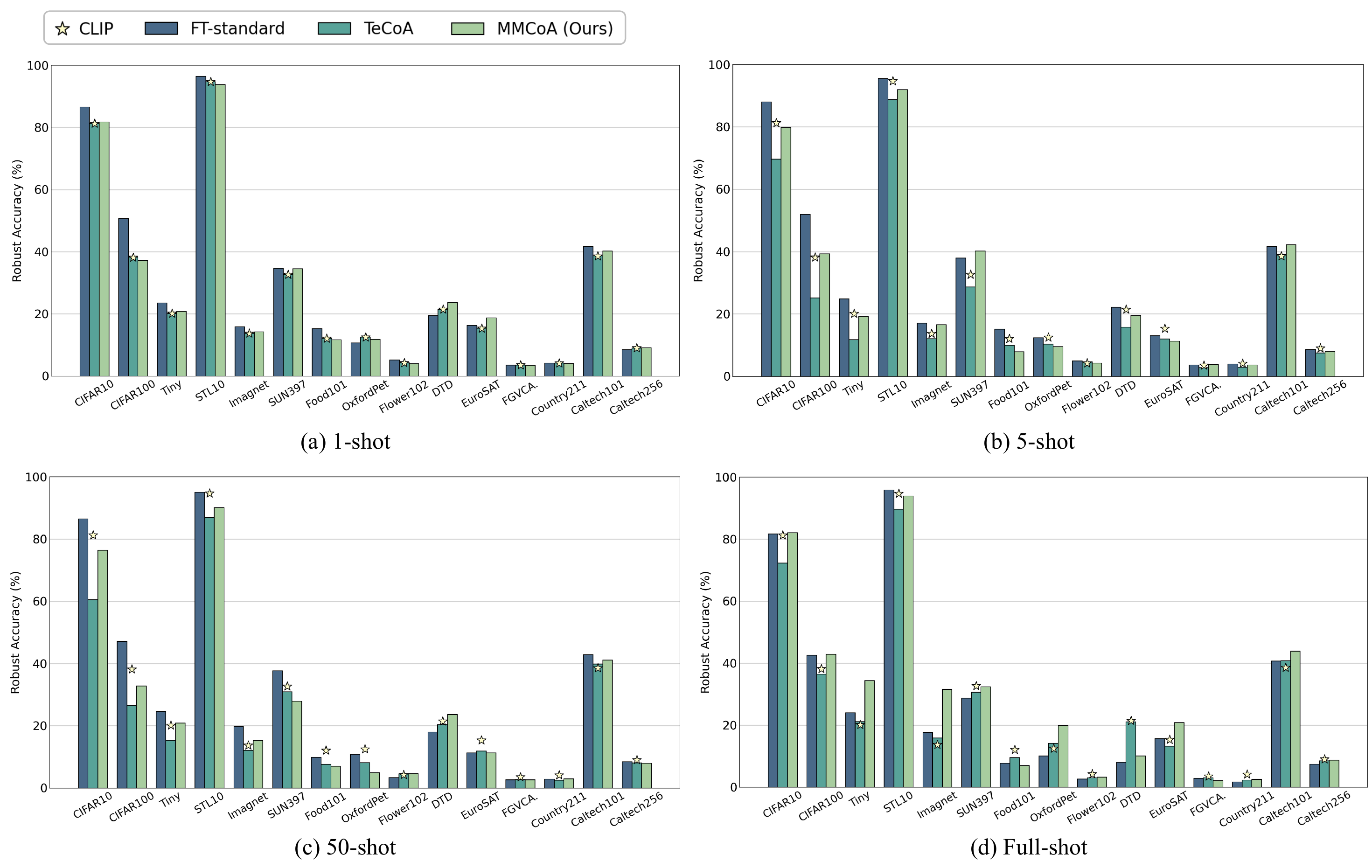}
  \caption{Out-of-distribution robust accuracies across 15 datasets under the \textbf{text-based BERT-Attack}. We fine-tune all methods on the ImageNet dataset with the few-shot setting and full-shot setting, and then test them on the remaining datasets.}
  \label{fig:ood_text}
\end{figure*}

\section{Experiments}
\label{sec:exp}

\subsection{Experimental Setup}

\subsubsection{Dataset}
\label{subsec:dataset}

To explore in-distribution adversarial robustness, we chose 4 datasets, namely CIFAR10~\cite{krizhevsky2009learning}, CIFAR100~\cite{krizhevsky2009learning}, TinyImageNet~\cite{deng2009imagenet}, and ImageNet~\cite{deng2009imagenet}.
For the exploration of out-of-distribution generalization in adversarial robustness, we chose 15 datasets. 
Specifically, we fine-tuned the models on ImageNet~\cite{deng2009imagenet} and evaluated their performance on the remaining 14 datasets. 
These datasets belong to two principal categories: generic object classification, which includes CIFAR10~\cite{krizhevsky2009learning}, CIFAR100~\cite{krizhevsky2009learning}, TinyImageNet (Tiny.)~\cite{deng2009imagenet}, STL10~\cite{coates2011analysis}, Caltech101~\cite{fei2004learning}, and Caltech256~\cite{griffin2007caltech}; and fine-grained classification, featuring OxfordPets~\cite{parkhi2012cats}, Flowers102~\cite{nilsback2008automated}, FGVCAircraft (FGVCA.)~\cite{maji2013fine}, Food101~\cite{bossard2014food}, EuroSAT~\cite{helber2018introducing}, DTD~\cite{cimpoi2014describing}, SUN397~\cite{xiao2010sun}, and Country211~\cite{radford2021learning}. 
Dataset details can be seen in the Appendix.

\subsubsection{Baselines} 
Given the limited research on the adversarial robustness of the CLIP model, we included three primary baselines:
\noindent \textbf{i) Zero-shot CLIP}~\cite{radford2021learning}, which uses the prompt ``This is a photo of a [CLS]" without fine-tuning;
\noindent \textbf{ii) FT-standard}~\cite{wang2024pre}, both the image and text encoders are fine-tuned on the clean dataset;
\noindent \textbf{iii) TeCoA}~\cite{mao2022understanding}, which employs text-guided contrastive adversarial training, leveraging additional information from text embeddings to correct the visual embeddings distorted by adversarial attacks.

\subsubsection{Implementation Details} 
We mainly adopted the CLIP-B/32 architecture of the image encoder as the backbone for all the methods used.
In our experiments, the hyperparameters $\gamma_1$ and $\gamma_2$ were set to 0.5, as specified in Equation~\eqref{func:total_loss}. 
We employed the Adam optimizer for model optimization, with beta coefficients of 0.9 and 0.98, a weight decay parameter of 0.2, and an initial learning rate of $1e-6$. 
Without additional specifications, for the adversarial training phase, we generated adversarial examples using a $10$-step PGD routine, and the step size $\alpha$ of $1/255$ with a perturbation limit set to an \( \mathcal{L}_{\infty} \) bound $\epsilon$ of $1/255$.
For the testing phase, we generated adversarial examples for images via a 100-step PGD method, each step exploring a step size $\alpha$ of $1/255$ with the perturbations bounded within an \( \mathcal{L}_{\infty} \) $\epsilon$ ball of $1/255$.
For the text modality attack, we followed the BERT-Attack~\cite{li2020bert} setup, using a maximum perturbation $\epsilon_t$ of 1 token and a selected word list length of 10.
For TeCoA, the learning rate was tuned between 1e-7 and 1e-5, with training durations ranging from 10 to 30 epochs, allowing for tailored optimization based on the method and dataset characteristics. The batch size was fixed at 256.
It should be noted that, unless otherwise specified, the default attack type was a white-box attack.

\begin{table*}[htbp]
\centering
\caption{Out-of-distribution \textbf{clean accuracies} across 15 datasets. Except for CLIP, we fine-tune all methods on ImageNet dataset with the few-shot setting (1-shot, 5-shot, and 50-shot) and full-shot setting, and then test them on the remaining datasets. Bold font denotes the best average accuracies.}
\label{tab:ood_clean}
\resizebox{\textwidth}{!}{
\begin{tabular}{lcccccccccccccccc}
\toprule
  & Source & \multicolumn{15}{c}{Target} \\
\cmidrule{2-17} 
Method & \rotatebox{90}{ImagNet} & \rotatebox{90}{CIFAR10} & \rotatebox{90}{CIFAR100} & \rotatebox{90}{Tiny.} & \rotatebox{90}{STL10} & \rotatebox{90}{SUN397} & \rotatebox{90}{Food101} & \rotatebox{90}{OxfordPets} & \rotatebox{90}{Flower102} & \rotatebox{90}{DTD} & \rotatebox{90}{EuroSAT} & \rotatebox{90}{FGVCA.} & \rotatebox{90}{Country211} & \rotatebox{90}{Caltech101} & \rotatebox{90}{Caltech256} & \rotatebox{90}{Average} \\
\midrule
CLIP & 59.13 & 88.57 & 62.22 & 57.90 & 97.15 & 57.64 & 83.84 & 87.38 & 65.70 & 40.21 & 38.49 & 19.98 & 15.22 & 84.91 & 87.38 & 63.05 \\
\midrule

\textit{1-shot} &  &  &  &  &  &  &  &  &  &  &  &  &  &  &  &  \\
\midrule
FT-standard & 63.87 & 90.03 & 67.23 & 61.89 & 97.26 & 62.42 & 84.43 & 89.32 & 65.99 & 42.61 & 47.33 & 20.85 & 15.90 & 86.69 & 83.01 & \textbf{65.26} \\
TeCoA & 59.13 & 88.60 & 62.32 & 57.91 & 97.15 & 57.64 & 83.35 & 87.38 & 65.70 & 40.21 & 38.49 & 19.98 & 15.23 & 84.91 & 82.05 & 62.67 \\
MMCoA & 58.80 & 89.87 & 62.39 & 57.69 & 97.20 & 57.98 & 83.61 & 86.56 & 65.25 & 39.95 & 40.74 & 19.80 & 15.28 & 84.27 & 81.83 & 62.75 \\
\midrule

\textit{5-shot} &  &  &  &  &  &  &  &  &  &  &  &  &  &  &  &  \\
\midrule
FT-standard & 65.50 & 91.43 & 68.57 & 62.34 & 96.69 & 62.27 & 83.56 & 88.66 & 63.56 & 39.04 & 45.29 & 19.14 & 16.03 & 85.64 & 82.35 & \textbf{64.67} \\
TeCoA & 51.92 & 84.05 & 46.54 & 39.45 & 94.44 & 54.53 & 72.53 & 78.74 & 57.52 & 32.50 & 22.91 & 16.65 & 9.85 & 82.47 & 78.49 & 54.84 \\
MMCoA & 57.37 & 87.00 & 55.97 & 48.96 & 94.18 & 61.28 & 72.15 & 82.97 & 57.16 & 34.47 & 30.58 & 15.00 & 12.88 & 84.58 & 79.81 & 58.29 \\
\midrule

\textit{50-shot} &  &  &  &  &  &  &  &  &  &  &  &  &  &  &  &  \\
\midrule
FT-standard & 61.05 & 88.87 & 61.90 & 57.46 & 96.06 & 58.18 & 75.46 & 83.51 & 54.32 & 37.71 & 22.01 & 11.37 & 13.12 & 84.04 & 80.30 & \textbf{59.02} \\
TeCoA & 49.89 & 74.64 & 41.69 & 46.41 & 91.67 & 56.11 & 55.66 & 75.47 & 49.55 & 31.91 & 19.60 & 14.64 & 9.16 & 82.60 & 76.08 & 51.67 \\
MMCoA & 60.07 & 78.63 & 48.12 & 58.67 & 90.80 & 55.93 & 60.21 & 76.81 & 46.28 & 32.39 & 23.51 & 7.53 & 9.77 & 81.02 & 76.59 & 53.76 \\
\midrule

\textit{Full-shot} &  &  &  &  &  &  &  &  &  &  &  &  &  &  &  &  \\
\midrule
FT-standard & 64.57 & 82.63 & 56.30 & 58.90 & 96.00 & 55.00 & 60.85 & 77.49 & 38.17 & 30.59 & 17.95 & 5.61 & 9.29 & 81.47 & 77.89 & 54.18 \\
TeCoA & 63.29 & 78.31 & 49.75 & 49.79 & 93.50 & 52.72 & 55.70 & 81.77 & 51.15 & 34.10 & 26.39 & 13.86 & 8.13 & 80.26 & 76.44 & 54.34 \\
MMCoA & 66.59 & 82.05 & 51.35 & 60.27 & 94.05 & 54.56 & 58.37 & 80.87 & 46.90 & 33.35 & 17.87 & 7.83 & 8.82 & 78.81 & 77.65 & \textbf{54.62} \\
\bottomrule
\end{tabular}
}
\end{table*}

\subsection{In-distribution Adversarial Robustness}
\label{subsec:iid}

For the in-distribution adversarial task, we aimed to examine the impact of various attacks on the CLIP model and the effectiveness of adversarial learning algorithms, particularly in scenarios where there is minimal distribution shift. 
To this end, we fine-tuned all methods using the training set of each dataset and then evaluated its performance on the test set from the same dataset.

As shown in Table~\ref{tab:iid}, we evaluated the in-distribution adversarial robustness across four datasets under three types of attacks.
When the task occurred with a small distribution shift, we obtained the following main findings:

\subsubsection{Multimodal adversarial training significantly enhances the adversarial robustness of both the image and text encoders}
compared with CLIP, MMCoA can outperform CLIP on four datasets with a large margin of around 45\%$\sim$75\%, 16\%$\sim$38\%, 28\%$\sim$72\% under image attack, text attack, and multimodal attack, respectively. 
Additionally, under multimodal attack, our MMCoA method significantly surpassed all baseline methods, demonstrating a substantial margin of improvement.

\subsubsection{The strength between image attack and text attack}
combining the results from Fig.~\ref{fig:intro} and Table~\ref{tab:iid}, we observed that 
(1) \textit{image attack has the potential to be more potent than text attack.} 
Specifically, a 100-step PGD image attack significantly compromised the adversarial robustness of CLIP. For instance, for CIFAR10, the accuracy declined from 88.57\% to 9.32\% under image attack, but under text attack, it only dropped to 79.74\%. 
Similarly, for CIFAR100, the accuracy decreased from 62.22\% to 4.59\% under image attack, whereas under text attack, it reduced to 37.24\%.
(2) \textit{As the number of categories in a dataset increased, text attacks became progressively stronger.} 
By comparing results across CIFAR10 (10 classes), CIFAR100 (100 classes), TinyImageNet (200 classes), and ImageNet (1000 classes), we observed a clear trend of decreasing classification accuracy under the text attacks.

\subsubsection{As the number of categories in a dataset increased, multimodal adversarial training was more effective than image adversarial training}
compared with TeCoA, a specialized adversarial fine-tuning algorithm designed to counter image attacks, our MMCoA may perform slightly worse on datasets with fewer categories, such as CIFAR10 and CIFAR100. However, as the number of categories in the experimental datasets increased, our MMCoA could even surpass TeCoA in defending image attack.

\subsubsection{Fine-tuning is a useful strategy for adapting VLMs to defend adversarial attacks} 
it is noteworthy that even when fine-tuning the CLIP model with a clean dataset or solely through image adversarial learning, its adversarial robustness was significantly enhanced across three types of attacks. 
This indicates that fine-tuning can adapt the model to a specific dataset, resulting in an overall improvement in the model's adversarial robustness within that dataset.

\begin{figure*}[ht]
  \centering
  \includegraphics[width=\textwidth]{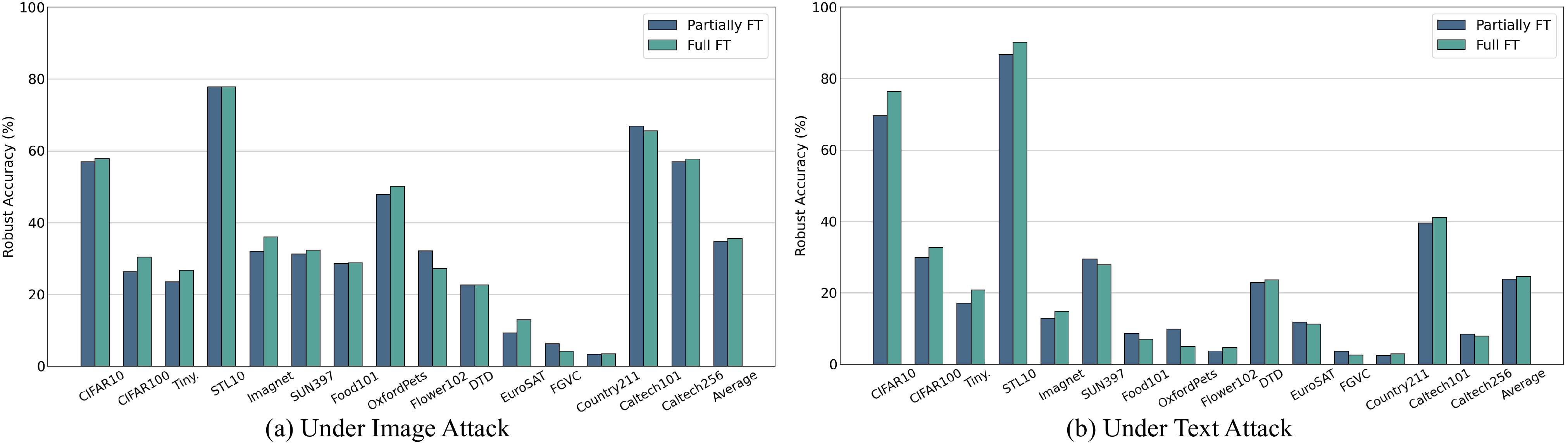}
  \caption{Exploration of the effect of the number of fine-tuned parameters with out-of-distribution generalization adversarial task on 15 datasets.}
  \label{fig:partially_ft_appendix}
\end{figure*}

\begin{figure*}[ht]
  \centering
  \includegraphics[width=\textwidth]{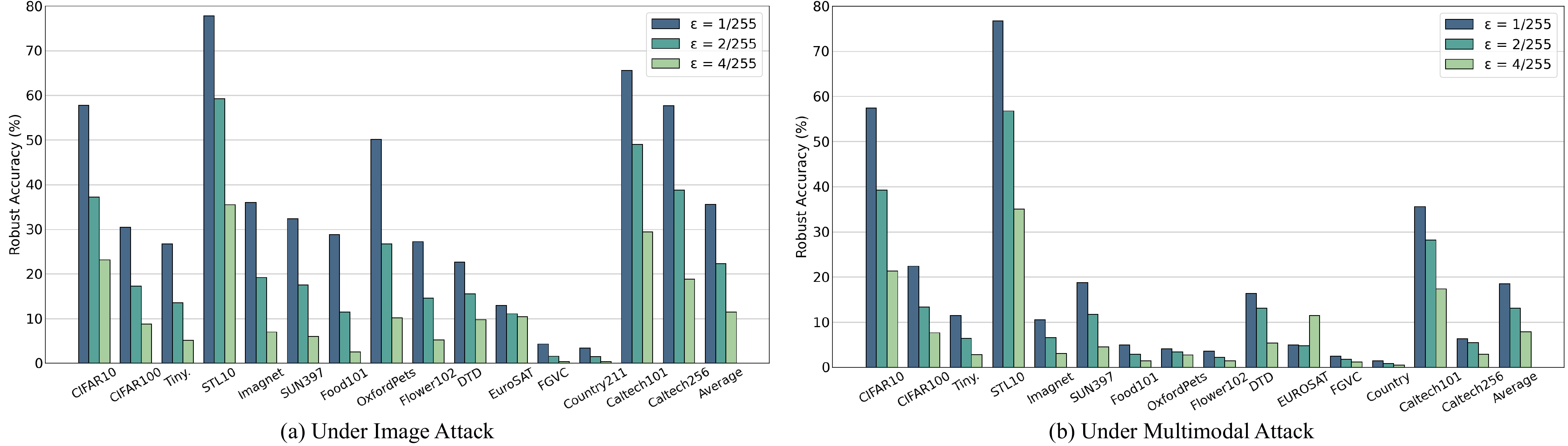}
  \caption{Exploration of the effect of the image attack and multimodal attack strength under different perturbation bounds with out-of-distribution generalization adversarial task on 15 datasets.}
  \label{fig:pb}
\end{figure*}

\subsection{Out-of-distribution Generalization Adversarial Robustness}

For the out-of-distribution generalization adversarial task, our objective was to explore the effectiveness of adversarial training on the CLIP model, when there is a significant distribution difference between the training and testing sets. 
To this end, we fine-tuned the model using ImageNet as the training dataset and then evaluated its performance across 14 distinct downstream datasets. 

It is important to highlight that findings 1), 2), and 4) detailed in Subsection~\ref{subsec:iid} remain valid even in most scenarios with significant distribution shifts. 
As demonstrated in Table~\ref{tab:ood} and Fig.~\ref{fig:ood_image}, our additional findings are:

\subsubsection{The larger the training set size, the stronger the model's adversarial robustness against multimodal and image attacks becomes} 
we note that the accuracies of FT-standard, TeCoA, and our MMCoA improved as the volume of training data increased, enhancing robust accuracy.

\subsubsection{Analysis of the adversarial robustness of baseline methods under multimodal and image attacks}
for FT-standard, fine-tuning with a clean dataset could enhance adversarial robustness, albeit marginally. 
In the case of TeCoA, its generalization capability was somewhat limited under a few-shot scenario. 
Nonetheless, with full-shot training, the improvement in adversarial robustness against multimodal and image attacks became significantly more pronounced.
While TeCoA slightly outperformed our MMCoA under image attacks, this suggests that image adversarial training may offer a slight advantage over multimodal adversarial training with more training examples. However, multimodal adversarial training proved to be more effective against multimodal and text attacks.

\subsubsection{Our MMCoA is effective in few-shot scenarios under multimodal and image attacks, where its generalization capability far surpasses that of the baselines} 
this demonstrates that our multimodal adversarial learning method can be significantly more efficient, making it an ideal choice for applications where data scarcity is a challenge. 
This balance between efficiency, effectiveness, and robust security underscores the potential of MMCoA as a frontier solution in the field of adversarial machine learning.

As shown in Fig.~\ref{fig:ood_text}, we also evaluated the out-of-distribution adversarial robustness across 15 datasets under text attack.
The main findings are as follows:

\subsubsection{FT-standard achieves state-of-the-art text adversarial robustness in the few-shot setting, but its accuracy decreases and its defensive capability against text attack diminishes as the number of samples increases}
this indicates that clean text is more effective in defending against text attack, and it can be achieved with only a small number of image-text pairs. 
However, in the full-shot scenario, overfitting to the source dataset may occur, potentially leading to a decrease in adversarial robustness.

\subsubsection{While aligning adversarial images with the corresponding correct texts, TeCoA proved ineffective under text attack} 
across all settings, TeCoA failed to surpass CLIP’s average robust accuracy. 
When facing minimal distribution shifts, image adversarial training paradigm could enhance the adversarial robustness under text attack. 
However, this enhancement was not present in scenarios involving large distribution shifts.

\subsubsection{MMCoA achieves state-of-the-art adversarial robustness in the full-shot scenario} 
it indicates that multimodal adversarial training requires a substantial number of samples to effectively counter text attack.

\begin{figure*}[ht]
  \centering
  \includegraphics[width=0.95\textwidth]{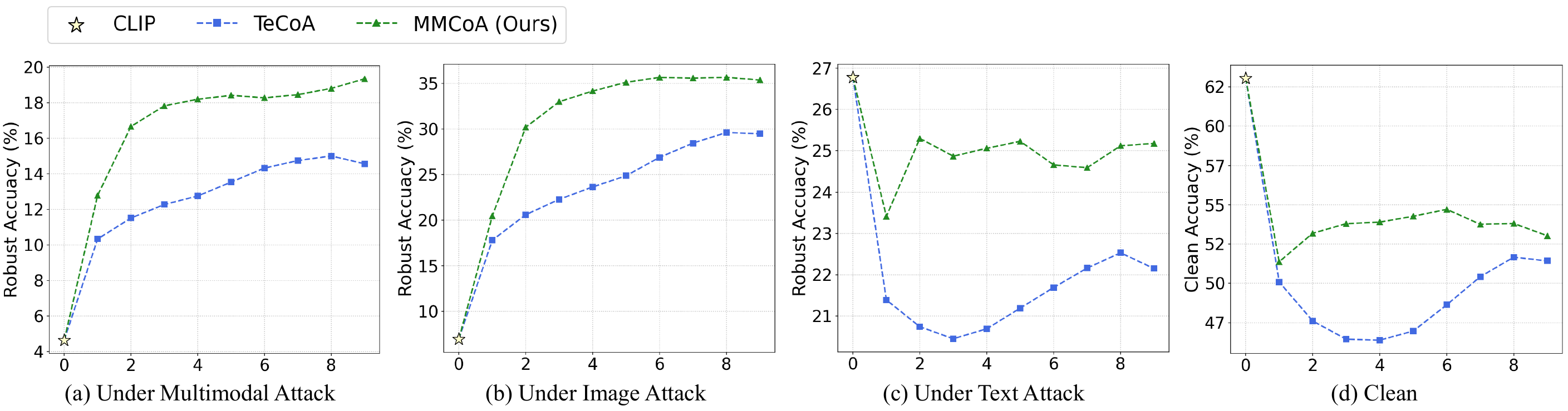}
  \caption{
   Effect of the number of iterations on averaged robust accuracy under different attacks and averaged clean accuracy in a 50-shot out-of-distribution generalization adversarial task across 15 datasets for CLIP, TeCoA, and our MMCoA.}
  \label{fig:iteration}
\end{figure*}

\begin{table*}[htbp]
\centering 
\caption{Ablation on each constraint loss across 4 datasets with the in-distribution adversarial task. Average accuracies under three types of attacks are reported. Bold denotes the best accuracy.}
\label{tab:ablation_loss_appendix}
\begin{tabular}{cc|cccc|cccc}
\midrule
\multirow{2}{*}{$\mathcal{L}(I'-T)$} & \multirow{2}{*}{$\mathcal{L}(I-T')$} & \multicolumn{4}{c|}{Multimodal Attack} & \multicolumn{4}{c}{Image Attack} \\
\cmidrule{3-10} 
 &  & CIFAR10 & CIFAR100 & Tiny. & Imagnet & CIFAR10 & CIFAR100 & Tiny. & ImageNet \\
\midrule
 &  & 11.65 & 2.42 & 2.18 & 1.80 & 9.32 & 4.59 & 1.93 & 0.79 \\
$\checkmark$ &  & 69.98 & 33.94 & 8.72 & 9.93 & 76.69 & 47.28 & 22.11 & 32.64 \\
 & $\checkmark$ & 28.49 & 10.62 & 1.72 & 8.01 & 28.85 & 9.90 & 0.92 & 3.14 \\
$\checkmark$ & $\checkmark$ & \textbf{84.52} & \textbf{55.13} & \textbf{39.19} & \textbf{30.02} & \textbf{84.55} & \textbf{59.50} & \textbf{53.97} & \textbf{46.58} \\
\midrule
\multirow{2}{*}{$\mathcal{L}(I'-T)$} & \multirow{2}{*}{$\mathcal{L}(I-T')$} & \multicolumn{4}{c|}{Text Attack} & \multicolumn{4}{c}{Clean} \\
\cmidrule{3-10} 
 &  & CIFAR10 & CIFAR100 & Tiny. & ImagNet & CIFAR10 & CIFAR100 & Tiny. & ImageNet \\
\midrule
 &  & 79.74 & 37.24 & 19.67 & 13.65 & 88.56 & 62.28 & 59.46 & 59.16 \\
$\checkmark$ &  & 86.75 & 47.96 & 16.34 & 13.42 & 93.42 & 70.23 & 49.98 & 53.47 \\
 & $\checkmark$ & \textbf{96.74} & \textbf{83.03} & \textbf{63.14} & \textbf{50.75} & \textbf{96.70} & \textbf{83.74} & 72.31 & 51.54 \\
$\checkmark$ & $\checkmark$ & 96.16 & 75.38 & 54.13 & 41.34 & 96.17 & 81.51 & \textbf{76.11} & \textbf{67.79} \\
\bottomrule
\end{tabular}
\end{table*}

\subsection{Clean Accuracy on Two Tasks} 

Previous work has primarily focused on studying the trade-off between robust accuracy and clean accuracy in the performance of adversarial training~\cite{wortsman2022robust}. 
However, in our study, we found that

\subsubsection{When facing minimal distribution shifts, robust accuracy and clean accuracy can be positively correlated}
as shown in Table~\ref{tab:iid}, regardless of whether clean or adversarial samples are incorporated into training, we observed that all methods could outperform CLIP in terms of both robust accuracy and clean accuracy. 
Notably, we did not observe this phenomenon where an increase in robust accuracy leads to a decline in clean accuracy.
Meanwhile, it is intuitive to see that training with clean samples allows FT-standard to achieve state-of-the-art clean accuracy on CIFAR10, CIFAR100, and TinyImageNet. 
However, on ImageNet, MMCoA reaches state-of-the-art in terms of clean accuracy.
This demonstrates that under minimal distribution shifts, adversarial training has the potential to simultaneously enhance both robust accuracy and clean accuracy, demonstrating exceptional performance.

\subsubsection{When facing large distribution shifts, the trade-off between robust accuracy under multimodal and image attacks and clean accuracy continues to hold} 
\begin{figure*}[ht]
  \centering
  \includegraphics[width=0.95\textwidth]{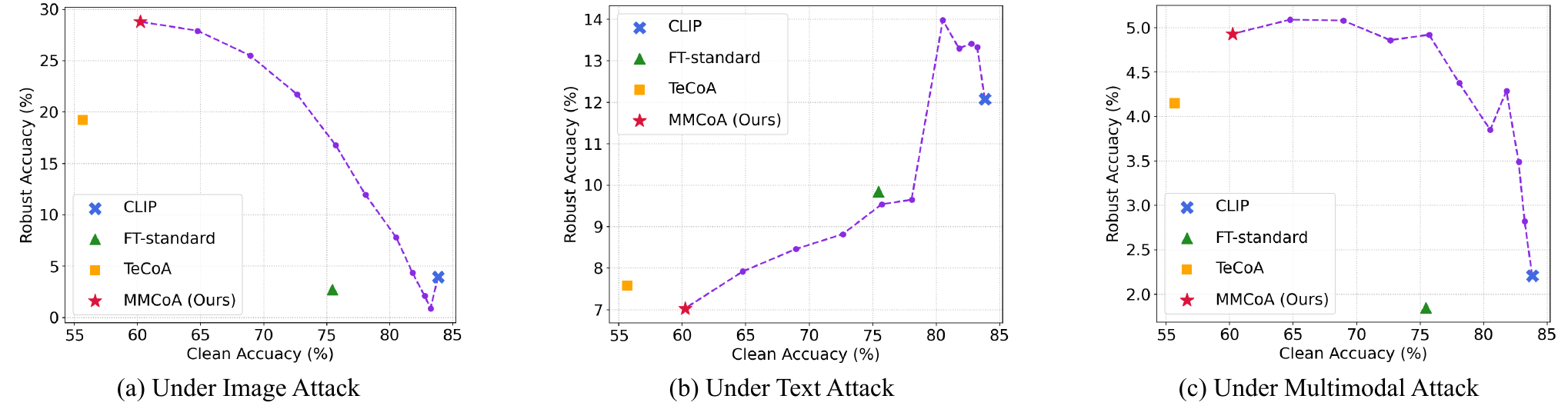}
  \caption{Relationship between the robust accuracy and the clean accuracy via weight interpolation. We test different methods on Food101 with 50-shot out-of-distribution generalization adversarial task under three types of attacks.}
  \label{fig:robustness-clean}
\end{figure*}

as illustrated in Table~\ref{tab:ood_clean}, Table~\ref{tab:ood}, and Fig.~\ref{fig:ood_image}, as the number of training examples increased, all methods demonstrated a trend whereby an increase in robust accuracy under multimodal and image attacks corresponds to a decrease in clean accuracy, thus indicating a clear trade-off. 
In addition to training examples, this observation aligns with the conclusions via weight interpolation, which is presented in Fig.~\ref{fig:robustness-clean}. 
Intriguingly, under text attack, the changes in text adversarial robustness closely paralleled the changes in clean accuracy.
Furthermore, FT-standard only surpassed CLIP in terms of average clean accuracy in the 1-shot and 5-shot settings. 
This enhanced performance with small sample sizes is also similar to FT-standard's behavior under text attack.

\begin{figure*}[ht]
  \centering
  \includegraphics[width=\textwidth]{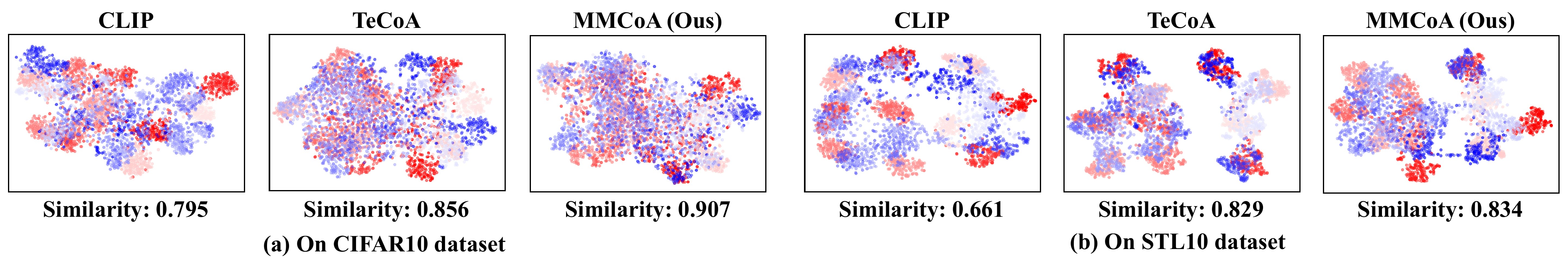}
  \caption{The t-SNE visualization on the CIFAR10 and STL10 with CLIP, TeCoA, and our MMCoA for out-of-distribution generalization adversarial task.
  The clean and adversarial image features extracted from 10 classes are shown in green and blue, respectively. Different shades of color represent different classes.}
  \label{fig:tsne}
\end{figure*}

\subsection{More Analysis}

\noindent \textbf{Effect of Each Constraint Loss.}
As shown in Table~\ref{tab:ablation_loss_appendix}, when minimal distribution shifts occur, the omission of the two contrastive losses significantly decreases robust accuracy across all three types of attacks.
Using only the text-supervised image adversarial loss resulted in a larger improvement in adversarial robustness against multimodal and image attacks. 
Conversely, employing only the image-supervised text adversarial loss notably enhanced both clean accuracy and robust accuracy against text attacks.
Within our MMCoA framework, the two types of losses work synergistically to improve performance, achieving state-of-the-art results in robust accuracy under multimodal and image attacks. 
However, under text attacks, our performance was weaker compared to the scenario where only the image-supervised text adversarial loss was used. 
This suggests that aligning adversarial images with clean text weakens the defense against text attacks.

Regarding clean accuracy, the image-supervised text adversarial loss contributed significantly more, especially in datasets with fewer categories, such as CIFAR10 and CIFAR100. 
For datasets with a larger number of categories, the interaction between the two losses proved to be more effective than using the image-supervised text adversarial loss alone.

\noindent \textbf{Effect of the Number of Iterations.}
As shown in Fig.~\ref{fig:iteration}, we evaluated CLIP, TeCoA, and our MMCoA with regard to the number of training iterations for robust accuracy under three types of attacks and clean accuracy.
In the 50-shot setting, TeCoA was originally trained over 10 epochs, reporting values for every epoch. MMCoA, being more challenging to fit, required additional epochs. 
Therefore, we demonstrated results over 28 epochs, reporting values every four epochs. 
Dataset-wise results are presented in the Appendix.

We observe that both TeCoA and MMCoA exhibit a trend where by robust accuracy initially increases and then stabilizes as the number of training iterations grows under multimodal and image attacks. 
The performance curves in Figs.~\ref{fig:iteration}(a) and \ref{fig:iteration}(b) for these two methods in these scenarios are notably similar, indicating comparable behavior in their ability to defend against these types of adversarial attacks.
However, in the case of text attacks and testing on clean datasets, the overall trend is a decrease followed by stabilization. 
The performance curves in Figs.~\ref{fig:iteration}(c) and \ref{fig:iteration}(d) follow a similar pattern, which, interestingly, is the opposite of the trends observed in Figs.~\ref{fig:iteration}(a) and \ref{fig:iteration}(b). 
This highlights that defending against text attacks is particularly challenging, especially when significant distribution shifts are present.

\noindent \textbf{Effect of Number of Fine-tuned Parameters.}
In this case, we tested the impact of only fine-tuning the image encoder, which is referred to as partial fine-tuning.
As shown in Fig.~\ref{fig:partially_ft_appendix}, we demonstrated a comparison of robust accuracy between partially fine-tuning and full fine-tuning under multimodal attack. 
It is evident that the more parameters are trained, the stronger the model's adversarial robustness. 

\noindent \textbf{Effect of Attack Strength.}
To investigate the impact of adversarial perturbation bounds on our method, we incrementally increased the perturbation bound for MMCoA among 1/255, 2/255, and 4/255 for both multimodal and image attacks.
As illustrated in Fig.~\ref{fig:pb}, our method exhibits a gradual decline in robust accuracy as the adversarial perturbation increases, with the decrease displaying a stepwise pattern. 

\noindent \textbf{Relationship between Robust and Clean Accuracy.}
As shown in Fig.~\ref{fig:robustness-clean}, manifested by large distribution shifts, MMCoA exhibits a trade-off between robust accuracy and clean accuracy under image and multimodal attacks, striking the most optimal balance compared to other methods. 
However, we observed a simultaneous decline in robust and clean accuracies under text attack, suggesting that training with adversarial text samples may not be a good choice when faced with significant distribution shifts. 
Instead, training with clean text proves can be more effective against text attack.

\noindent \textbf{T-SNE Visualizations.}
As shown in Fig.~\ref{fig:tsne}, we visualized the clean and adversarial image features learned by CLIP, TeCoA, and MMCoA on two datasets.
Qualitatively, we could observe that MMCoA achieved a higher degree of mix between clean and adversarial image features compared to the baseline methods. 
Quantitatively, we calculated the cosine similarity between the clean and adversarial features for each image and computed the average value across all samples.
We found that our MMCoA exhibited the highest similarity between the clean and adversarial features for each image.
The closer alignment between clean and adversarial features further implies that MMCoA mitigates the impact of adversarial perturbations, thereby improving the adversarial robustness of CLIP.

\begin{table}[ht]
\caption{In-disribution robust accuracies on 3 datasets and comparison of out-of-distribution robust average accuracies on 13 datasets under Co-Attack.}
\label{tab:coattack}
\centering
\resizebox{0.5 \textwidth}{!}{
\begin{tabular}{lccc|ccc}
\toprule
 & \multicolumn{3}{c|}{In-disribution} & \multicolumn{3}{c}{Out-of-distribution} \\
\cmidrule(r){2-7} 
Method & CIFAR10 & CIFAR100 & Tiny. & 1-shot & 5-shot & 50-shot \\
\midrule
CLIP & 39.72 & 9.17 & 0.46 & 13.47  & 13.47 & 13.47 \\
FT-standard & 58.48 & 12.49 & 1.00 & 13.99 & 15.36 & 17.70 \\
TeCoA & 85.58 & 48.13 & 17.94 & 13.53 & 15.16 & 15.42 \\
MMCoA & \textbf{87.96} & \textbf{60.40} & \textbf{22.88} & \textbf{14.57} & \textbf{20.74} & \textbf{22.66} \\
\bottomrule
\end{tabular}
}
\end{table}

\noindent \textbf{Additional Results on More Multimodal Attacks.}
We conducted evaluation experiments under Co-Attack~\cite{zhang2022towards},
which is a collective multimodal attack that models cross-modal interactions. 
To attack the CLIP model, Co-Attack first generates adversarial text using BERT-Attack, and then generates adversarial images by maximizing the KL divergence between the adversarial text embedding and the adversarial image embedding, as well as the KL divergence between the adversarial and clean image embeddings.
In these experiments, we tested the models trained in our previous experiments (under both in-distribution and out-of-distribution settings) against the 100-step Co-Attack. 
As shown in Table~\ref{tab:coattack}, our MMCoA method demonstrates superior effectiveness in defending against multimodal attacks.

\noindent \textbf{Additional Results on Image-text Retrieval Task.}
We evaluated TeCoA and our MMCoA method, trained on the ImageNet-1K dataset, on the MSCOCO dataset~\cite{lin2015microsoftcococommonobjects} under Co-Attack~\cite{zhang2022towards} for the image-to-text retrieval (TR) and text-to-image retrieval (IR). 
The R@K (Recall at K) represents the percentage of cases where the correct result is ranked as the top $K$ result. 
The maximum perturbation $\epsilon_i$ was set to $2/255$, the step size was set to $1.25$, and the number of iterations was set to $10$.
As shown in Table~\ref{tab:retrival}, MMCoA exhibits superior effectiveness in defending against multimodal attacks in the image-text retrieval task, indicating a stronger image-text alignment capability. 
Notably, the reported value is robust accuracy and the step size used in this experiment follows the settings defined in the Co-Attack project\footnote{https://github.com/adversarial-for-goodness/Co-Attack.git}.

\section{Comprehensive Summary}
\label{sec:important}

In this section, we provide a summary of adversarial robustness under three types of attacks across two tasks, as well as the clean accuracy of each method on clean samples. 
Firstly, we extend the concept of distribution shifts between datasets to include variations in textual categories, in addition to the distribution shifts among images, such as differences in category types and the number of categories. Therefore, in this paper, significant distribution shifts refer not only to substantial variations in image distributions but also to considerable disparities in text distributions.

\begin{table}[htbp]
\caption{Robust accuracy comparison on MSCOCO dataset with image-text retrieval task under Co-Attack.}
\label{tab:retrival}
\centering
\begin{tabular}{lccc|ccc}
\toprule
\multicolumn{1}{c}{} & \multicolumn{3}{c}{TR} & \multicolumn{3}{c}{IR} \\
\cmidrule(lr){2-4} \cmidrule(lr){5-7}
Method & R@1 & R@5 & R@10 & R@1 & R@5 & R@10 \\
\midrule
CLIP & 3.94 & 10.34 & 15.42 & 1.73 & 5.23 & 7.99 \\
TeCoA & 10.94 & 26.14 & 35.76 & \textbf{7.40} & 18.46 & 26.33 \\
MMCoA & \textbf{13.22} & \textbf{29.94} & \textbf{39.12} & 7.39 & \textbf{18.98} & \textbf{26.55} \\
\bottomrule
\end{tabular}
\end{table}

\subsection{In-distribution Adversarial Task}

\noindent \textbf{Adversarial Robustness.} 
With minimal distribution shifts, by fine-tuning the CLIP model with a clean dataset, solely through image adversarial contrastive training, or through our multimodal adversarial contrastive training, all methods enhanced CLIP's adversarial robustness under the three types of attacks.
Adversarial training was particularly effective under multimodal and image attacks. 
Additionally, our multimodal contrastive adversarial training (MMCoA) demonstrated superior adversarial robustness in more complex datasets, such as those with a greater number of categories, across all three types of attacks.

\noindent \textbf{Clean Accuracy.}
When facing minimal distribution shifts, there is often a positive correlation between robust accuracy and clean accuracy, indicating that the traditional trade-off between robustness and performance may not hold under these conditions. 
Moreover, fine-tuning the CLIP model with a clean dataset yielded optimal clean accuracy on CIFAR10, CIFAR100, and TinyImageNet.
However, for ImageNet, the MMCoA method achieved state-of-the-art clean accuracy, underscoring the potential of adversarial learning to enhance clean accuracy when fine-tuning specific datasets with adversarial examples on large-scale multimodal models like CLIP.

\subsection{Out-of-distribution Generalization Adversarial Task}

\noindent \textbf{Adversarial Robustness.}
With large distribution shifts, all three methods still enhanced CLIP's adversarial robustness under two types of attacks in both few-shot and full-shot settings, \textit{i.e.,} multimodal and image attacks, achieving optimal performance in the full-shot scenario. 
Moreover, our multimodal contrastive adversarial (MMCoA) training effectively balanced efficiency, effectiveness, and robustness.
However, for text attack, none of the methods significantly enhanced CLIP's robustness, indicating that defense strategies against text attack require further exploration. 
It is our hope that this study can serve as a foundation for future research to develop more effective methods for defending against text-based adversarial attacks.

\noindent \textbf{Clean Accuracy.}
When facing large distribution shifts, there is a clear trade-off between robust accuracy under multimodal and image attacks and clean accuracy, consistent with previous research that adversarial training can compromise performance on clean datasets. 
Interestingly, the variations in robust accuracy under text attack showed some similarity to changes in clean accuracy.

\section{Limitations and Conclusions}
\label{sec:conclu}

In this paper, we have conducted the first comprehensive study aimed at enhancing the adversarial robustness of VLMs against attacks targeting visual, textual, and multimodal inputs.
To fill the gap in multimodal defense, we have introduced a multimodal contrastive adversarial training framework to effectively enhance the adversarial robustness of both the image and text encoders within CLIP.
Our extensive experiments and analyses have highlighted the performance and characteristics of various adversarial defense methods under different distribution shifts and dataset complexities across the three attack types.
In particular, we have found that text adversarial training can help mitigate the trade-off between clean accuracy and robustness, and that multimodal adversarial training boosts the adversarial robustness of few-shot fine-tuning under image attack.
It is our hope that these findings will pave the way for new robust strategies to strengthen the security frameworks of VLMs, driving further advancements that can be quickly translated into real-world applications.
Importantly, we have found that defending against text attacks remains a challenge under large distribution shifts in text domain. 
We leave this topic for future exploration.





\bibliographystyle{IEEEtran}
\bibliography{main}

\appendix

In the Appendix, we give complementary descriptions of datasets and provide more comprehensive experimental evaluations.

\subsection{Dataset}
\label{sec:DD}

We evaluate the adversarial robustness on 15 datasets, which can fall into two categories: generic object classification and fine-grained classification. We first introduce \textit{generic object classification} datasets as follows.

\noindent 
\textbf{CIFAR10}~\cite{krizhevsky2009learning} originates from the Tiny Images dataset, featuring 60,000 color images with dimensions of $32\times32$. Each image is uniquely classified into one of ten distinct classes.

\noindent \textbf{CIFAR100}~\cite{krizhevsky2009learning} also derived from the Tiny Images, includes 60,000 color images, each measuring 32x32 pixels. CIFAR100 is organized into 100 classes, which are further aggregated into 20 superclasses for a structured classification framework.

\noindent \textbf{TinyImageNet}~\cite{deng2009imagenet} offers a compact version of the ImageNet, containing 100,000 images across 200 classes, with each class providing 500 training images, alongside 50 validation and 50 test images, all resized to 64×64 pixels.

\noindent \textbf{STL10}~\cite{coates2011analysis} is a subset of ImageNet~\cite{deng2009imagenet}. It includes 13,000 color images with a resolution of 96×96, representing 10 object classes.

\noindent \textbf{Caltech101}~\cite{fei2004learning} contains images from 101 object categories and a background category that contains the images not from the 101 object categories. For each object category, there are about 40 to 800 images, while most classes have about 50 images. The resolution of the image is roughly about 300×200 pixels.

\noindent \textbf{Caltech256}~\cite{griffin2007caltech} an extension of Caltech101~\cite{fei2004learning}, is an object recognition dataset with 30,607 images of varying sizes across 257 categories (256 object classes plus one clutter class), ensuring a minimum of 80 images per class.

\noindent \textbf{ImageNet}~\cite{deng2009imagenet}  contains 14,197,122 annotated images according to the WordNet hierarchy, which is a large-scale image database. This expansive collection spans roughly 22,000 categories, encompassing a diverse array of subjects including various animals, plants, geographical locations, and common everyday items.

Then we introduce the remaining \textit{fine-grained classification }datasets as follows.

\noindent 
\textbf{OxfordPets}~\cite{parkhi2012cats} is comprised of a dataset featuring 37 categories of pets, each represented by approximately 200 images. These images exhibit a broad variety in scale, pose, and lighting conditions. Each image is meticulously annotated with breed information, a region of interest for the head, and pixel-level trimap segmentation for detailed analysis.

\noindent \textbf{Flower102}~\cite{nilsback2008automated}  introduces a dataset of 102 flower categories, selected for their common occurrence in the United Kingdom. Each category contains between 40 and 258 images. Detailed information on the categories and the specific number of images per class is accessible through the category statistics page.

\noindent \textbf{FGVCAircraft}~\cite{maji2013fine}  encompasses 10,200 images across 102 different aircraft model variants, mainly featuring airplanes. Each image highlights the main aircraft within a precise bounding box and is classified using a hierarchical airplane model label, organized across four levels of hierarchy.

\noindent \textbf{Food101}~\cite{bossard2014food} includes 101 food categories, each with 750 training images and 250 test images, totaling 101,000 images. The test images have undergone manual cleanup to ensure label accuracy, whereas the training set may include some level of noise. 

\noindent \textbf{EuroSAT}~\cite{helber2018introducing} is a dataset and deep learning benchmark for land use and land cover classification. The dataset is based on Sentinel-2 satellite images covering 13 spectral bands and consisting of 10 classes with in total of 27,000 labeled and geo-referenced images.

\noindent \textbf{DTD}~\cite{cimpoi2014describing} offers 5,640 images of natural textures, annotated with attributes that reflect the perceptual properties of textures as perceived by humans. Organized by a list of 47 perceptually inspired terms, DTD includes 120 images for each category. Image sizes vary, and each predominantly showcases the texture attribute it represents. Sourced from Google and Flickr, these images were annotated via Amazon Mechanical Turk through multiple rounds.

\noindent \textbf{SUN397}~\cite{xiao2010sun} contains 899 categories and 130,519 images. There are 397 well-sampled categories to evaluate numerous algorithms for scene recognition.

\noindent \textbf{Country211}~\cite{radford2021learning} is a dataset released by OpenAI, designed to assess the geolocation capability of visual representations. It filters the YFCC100m~\cite{thomee2016yfcc100m} to find 211 countries that have at least 300 photos with GPS coordinates. OpenAI built a balanced dataset with 211 categories, by sampling 200 photos for training and 100 photos for testing, for each country.
\begin{figure*}[htbp]
  \centering
  \includegraphics[width=0.82\textwidth]{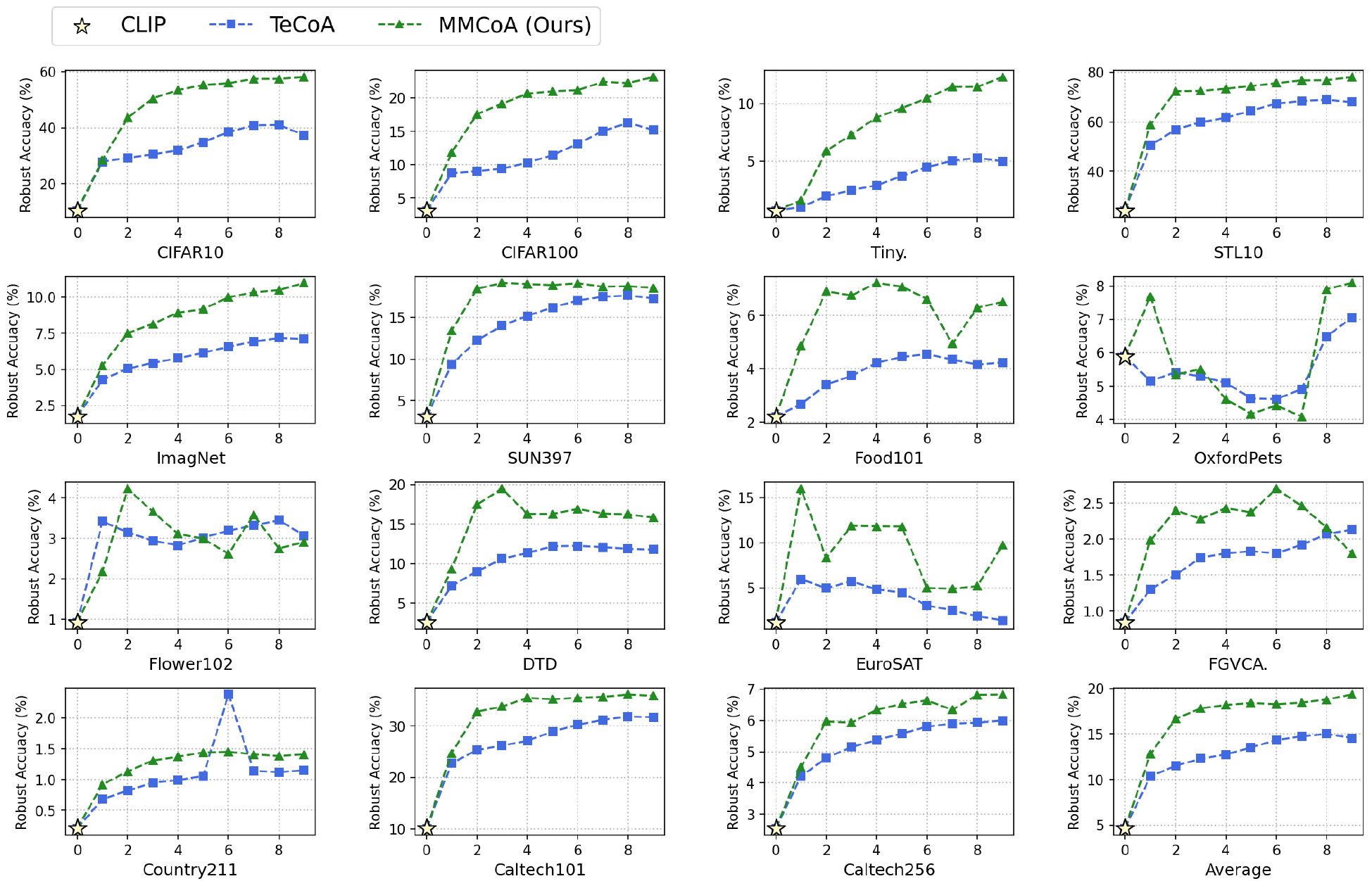}
  \caption{Effect of the number of iterations on 50-shot out-of-distribution generalization adversarial task for CLIP, TeCoA, and our MMCoA across 15 datasets under \textbf{multimodal attack}.}
  \label{fig:iteration_multimodal}
\end{figure*}

\begin{figure*}[htbp]
  \centering
  \includegraphics[width=0.82\textwidth]{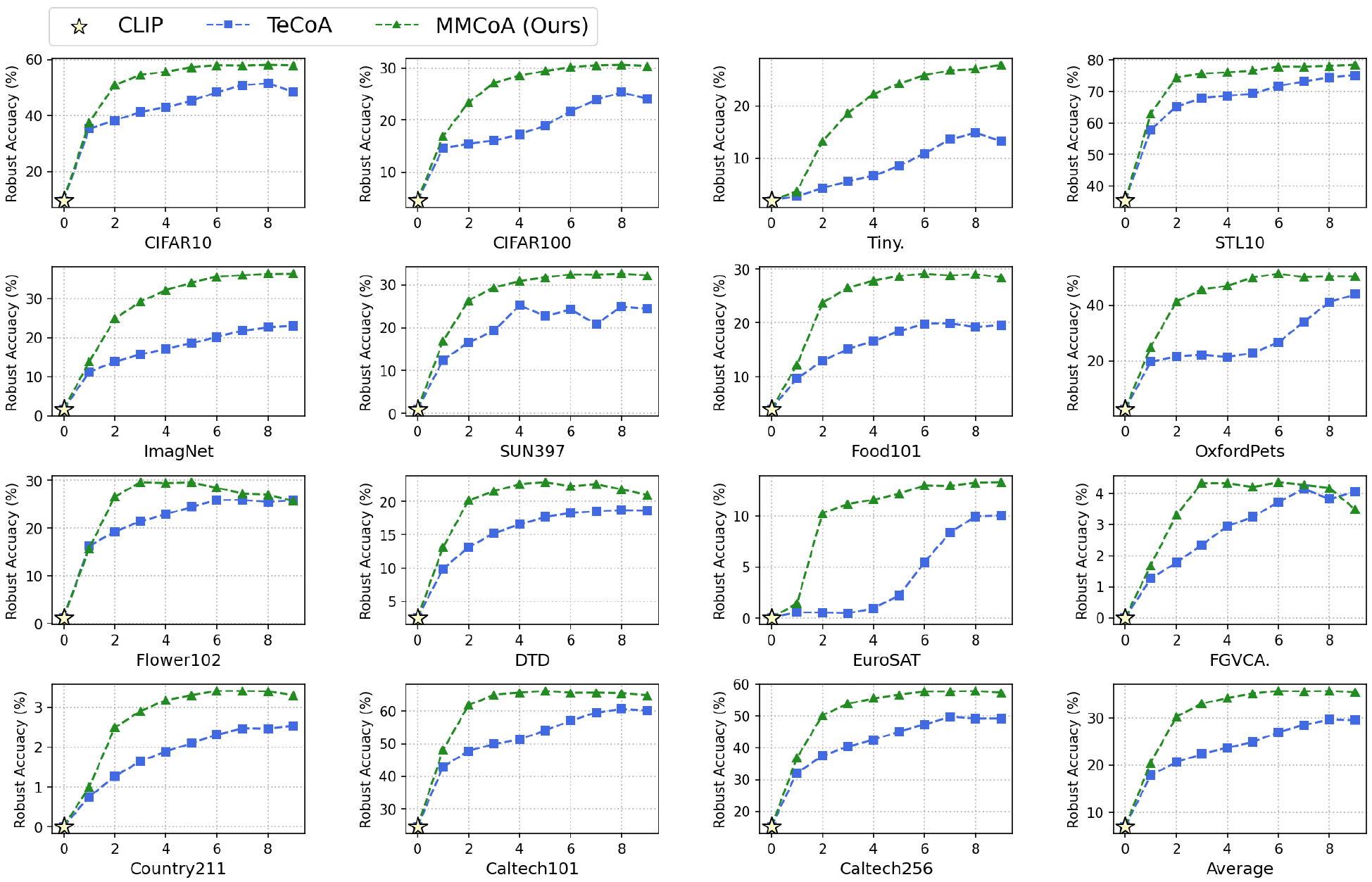}
  \caption{Effect of the number of iterations on 50-shot out-of-distribution generalization adversarial task for CLIP, TeCoA, and our MMCoA across 15 datasets under \textbf{image attack}.}
  \label{fig:iteration_image}
\end{figure*}

\begin{figure*}[htbp]
  \centering
  \includegraphics[width=0.82\textwidth]{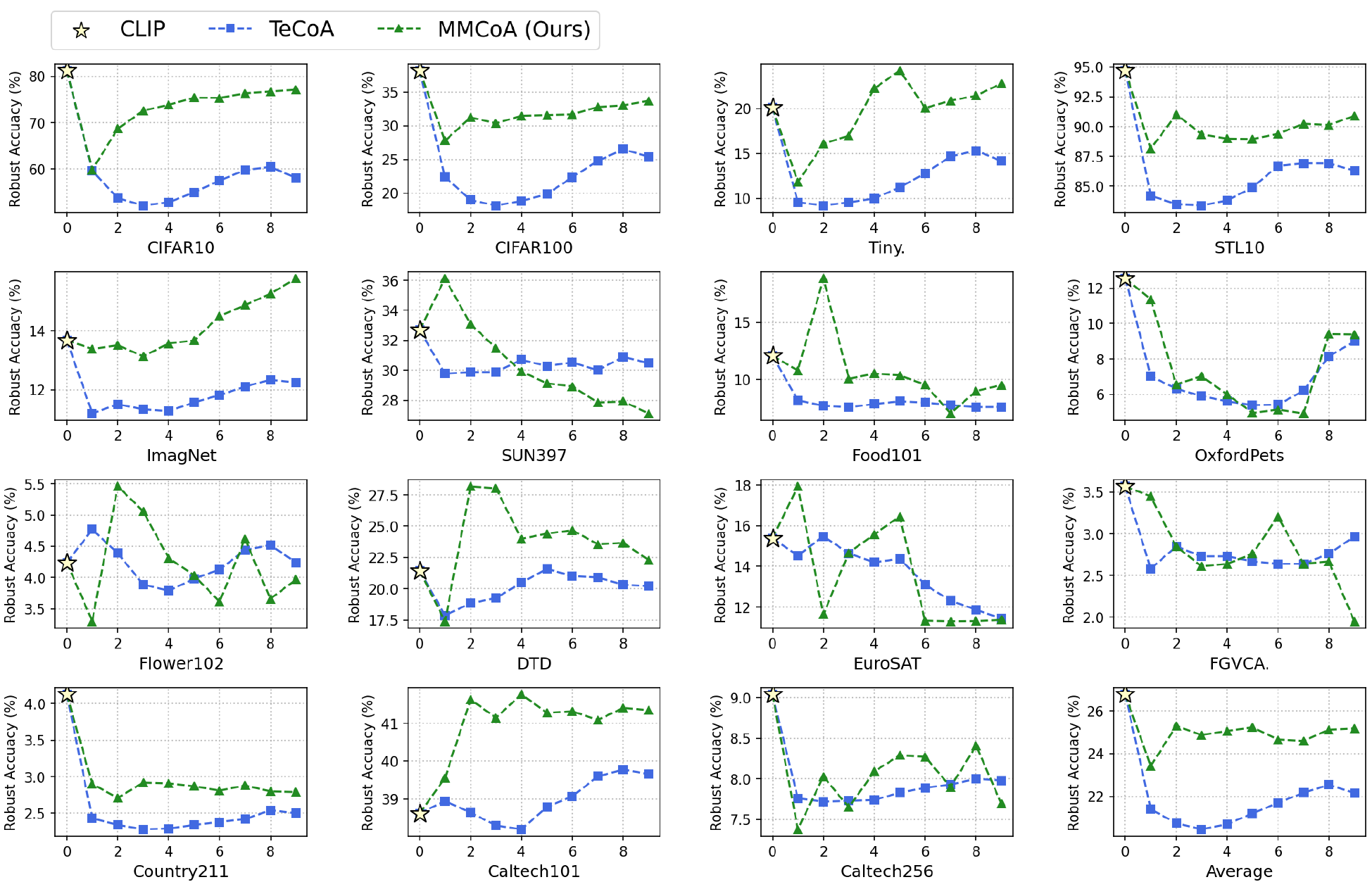}
  \caption{Effect of the number of iterations on 50-shot out-of-distribution generalization adversarial task for CLIP, TeCoA, and our MMCoA across 15 datasets under \textbf{text attack}.}
  \label{fig:iteration_text}
\end{figure*}

\begin{figure*}[htbp]
  \centering
  \includegraphics[width=0.82\textwidth]{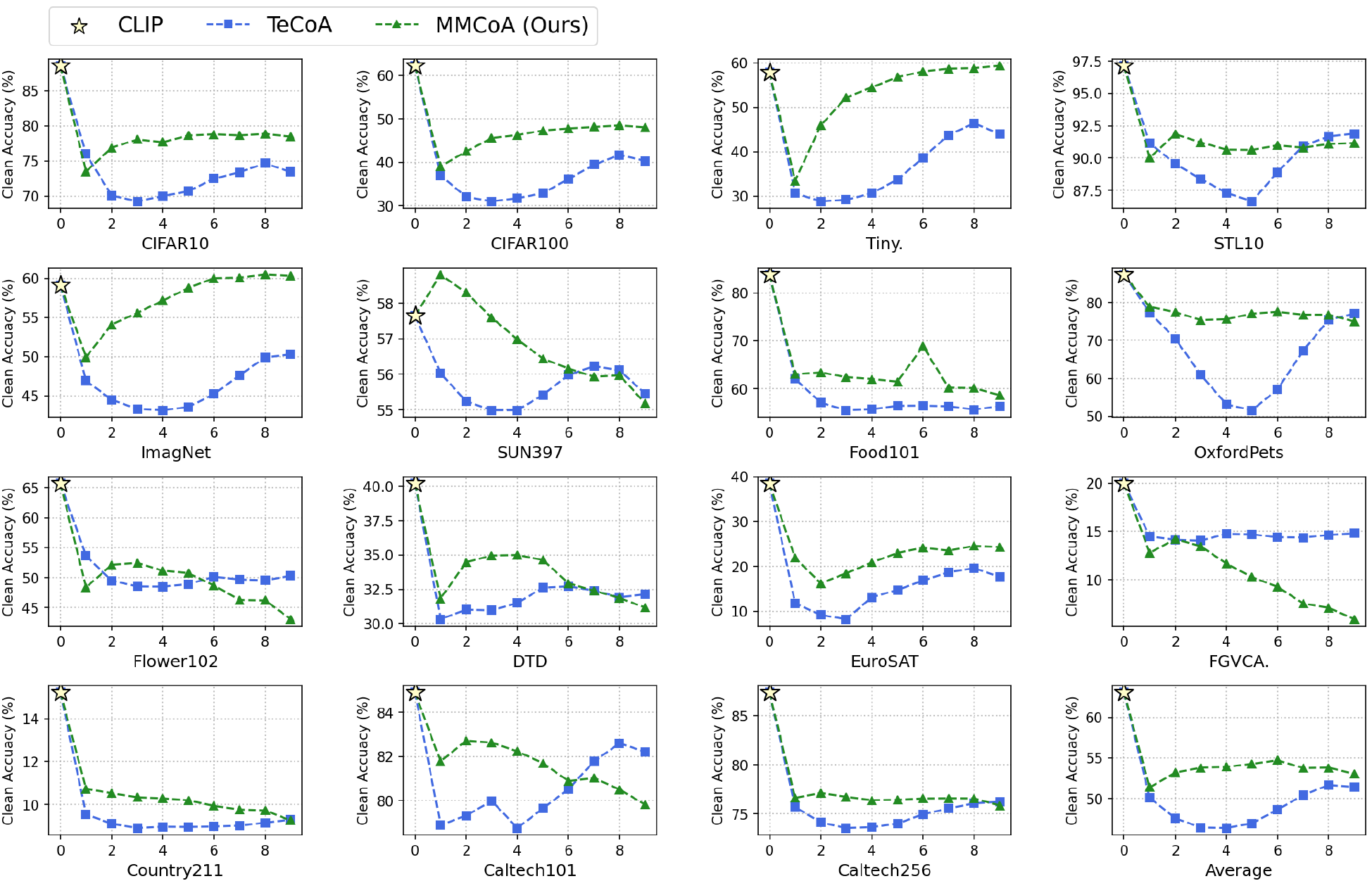}
  \caption{Effect of the number of iterations on 50-shot out-of-distribution generalization adversarial task for CLIP, TeCoA, and our MMCoA across 15 datasets with \textbf{clean accuracy}.}
  \label{fig:iteration_clean}
\end{figure*}

\subsection{Additional Results on More Methods}

In addition to full fine-tuning and partial fine-tuning, we also compared the current popular prompt-tuning methods for adversarial defense. We conducted experiments with AdvPT~\cite{zhang2023adversarial} and VP~\cite{chen2023visual}, and as shown in Table~\ref{tab:prompt}, while these methods improve CLIP's robustness, they are not as effective as TeCoA or our method. Therefore, although prompt tuning introduces fewer parameters, its adversarial robustness remains limited and requires further improvement.

\begin{table}[htbp]
\centering
\caption{IID robust accuracies comparison with VP and APT under image attack (I.A.), text attack (T.A.) and multimodal attack (MM.A.).}
\label{tab:prompt}
\begin{tabular}{lccc|ccc}
\toprule
 & \multicolumn{3}{c|}{CIFAR10} & \multicolumn{3}{c}{CIFAR100} \\
\cmidrule(r){2-7} 
Method & I.A. & T.A. & MM.A. & I.A. & T.A. & MM.A. \\
\midrule
AdvPT & 66.70 & -   & -   & 44.20  & -    & -   \\
VP & 74.52 & 81.26 & 66.79 & 45.35 & 38.22 & 28.37 \\
TeCoA & \textbf{85.06} & 92.04 & 79.53 & \textbf{60.71} & 56.96 & 44.34 \\
MMCoA  & 84.55 & \textbf{96.16} & \textbf{84.52} & 59.50 & \textbf{75.38} & \textbf{55.13}  \\
\bottomrule
\end{tabular}
\end{table}

\subsection{Effect of the Number of Iterations.}
As shown in Figs.~\ref{fig:iteration_multimodal}$\sim$\ref{fig:iteration_clean}, we evaluated CLIP, TeCoA, and our method with regard to the number of training iterations for robust accuracy under three types of attacks and clean accuracy.
In the 50-shot setting, TeCoA originally trained for 10 epochs, reporting values for every epoch. 
MMCoA, being more challenging to fit, required additional epochs. 
Therefore, we demonstrated results over 28 epochs, reporting values every four epochs.

We observed that both TeCoA and MMCoA exhibited a trend where robust accuracy initially increased and then stabilized as the number of training iterations grew under multimodal and image attacks. 
The performance curves for these two methods in these scenarios are notably similar, indicating comparable behavior in their ability to defend against these types of adversarial attacks.
However, in the case of text attacks and testing on clean datasets, only a few datasets showed an upward trend in performance. 
Notably, aside from ImageNet, which served as the source dataset, Caltech101 and TinyImageNet also exhibited improvements. 
This is likely due to the relatively minor distributional differences between these datasets and ImageNet, making the models more adaptable to these target distributions. 
The performance curves in these cases also showed a similar pattern.
It implies that distribution similarity plays a crucial role in adversarial training, and further investigation is needed to enhance robustness in more diverse and challenging text-based adversarial settings.

\subsection{Time Cost}

Table~\ref{tab:time} presents the training time on NVIDIA A6000 GPU. For CIFAR10, CIFAR100, and TinyImageNet, we used 2 GPUs, while for ImageNet, we used 4 GPUs.

\begin{table}[htbp]
\centering
\caption{Time cost.}
\label{tab:time}
\begin{tabular}{ccc}
\toprule
Dataset      & Training Time (per epoch) & Batch Size \\
\midrule
\multicolumn{3}{l}{\textit{In-distribution setting.}}                   \\
\midrule
CIFAR10      & 40min                     & 128        \\
CIFAR100     & 45min                     & 128        \\
TinyImageNet & 1.30h                     & 128        \\
ImageNet     & 2.25h                     & 256        \\
\midrule
\multicolumn{3}{l}{\textit{Out-of-distribution setting.}}               \\
\midrule
1-shot       & 3min                      & 256        \\
5-shot       & 9min                      & 256        \\
50-shot      & 1.15h                     & 256     \\
\bottomrule
\end{tabular}
\end{table}

\end{document}